\documentclass[9pt, a4paper, twocolumn]{arxiv}

\usepackage[all]{hypcap}
\usepackage[comma,numbers,sort,compress]{natbib}
\usepackage[capitalize,noabbrev]{cleveref}
\usepackage{graphicx}
\usepackage{subfigure}
\usepackage{float}
\usepackage{wrapfig}
\usepackage{placeins}
\usepackage[hang,flushmargin]{footmisc}

\usepackage{booktabs}
\usepackage{colortbl}
\usepackage{multirow}
\usepackage{tabularx}

\usepackage{pifont}

\newcommand{\betterV}[1]{\cellcolor{red!25}\textbf{#1}}

\newcommand{\myblue}[1]{\textbf{\textcolor[rgb]{0.4,0.4,0.9}{#1}}}

\definecolor{orange1}{RGB}{198, 95, 16}
\newcommand{\myorangetext}[1]{\textcolor{orange1}{#1}}

\definecolor{blue1}{RGB}{46, 84, 161}
\newcommand{\mybluetext}[1]{\textcolor{blue1}{#1}}

\definecolor{green1}{RGB}{88, 142, 49}
\newcommand{\mygreentext}[1]{\textcolor{green1}{#1}}

\newcommand{\fstP}[1]{\cellcolor{red!40}\textbf{#1}}
\newcommand{\sndP}[1]{\cellcolor{orange!40}{#1}}
\newcommand{\tidP}[1]{\cellcolor{yellow!40}{#1}}

\usepackage{amsmath}
\usepackage{amssymb}
\usepackage{mathtools}
\usepackage{amsthm}

\usepackage{algorithm}
\usepackage{algpseudocode}

\usepackage{microtype}
\usepackage{setspace}
\usepackage{xcolor}
\usepackage[most]{tcolorbox}

\usepackage{enumitem}

\usepackage{listings}
\usepackage{color}

\definecolor{sb_blue}{RGB}{65,105,225}
\definecolor{sb_orange}{RGB}{255,140,0}
\definecolor{gray}{RGB}{128,128,128}

\lstdefinestyle{mystyle}{
    language=Python,
    xleftmargin=5.0ex,
    basicstyle=\footnotesize\ttfamily\linespread{4},
    backgroundcolor=\color{gray!10},
    commentstyle=\color{gray},
    alsoletter={<>-0123456789},
    keywordstyle=\color{sb_blue},
    ndkeywords={nn, F, torch, partial},
    ndkeywordstyle=\color{sb_orange},
    emph={import, def, return, if, else},
    emphstyle=\bfseries\color{sb_blue},
    numberstyle=\footnotesize\ttfamily\color{gray},
    stringstyle=\color{sb_blue},
    breakatwhitespace=false,
    breaklines=true,
    keepspaces=true,
    numbers=left,
    numbersep=5pt,
    showspaces=false,
    showstringspaces=false,
    showtabs=false,
    tabsize=2
}
\definecolor{lightblue}{rgb}{0.22,0.45,0.70}
\newcommand{\Appendix}{\textcolor{mylinkcolor}{Appendix}}
\newcommand{\Figure}{\textcolor{mylinkcolor}{Figure}}
\newcommand{\Section}{\textcolor{mylinkcolor}{Section}}
\newcommand{\Table}{\textcolor{mylinkcolor}{Table}}
\newcommand{\Equation}{\textcolor{mylinkcolor}{Equation}}

\title{Exploration Matters for Escaping the Blur Trap \\ in 3D Gaussian Splatting}
\shorttitle{Exploration Matters for Escaping the Blur Trap in 3D Gaussian Splatting}

\author[1]{Chengbo Wang}
\author[2]{Guozheng Ma}
\author[1]{Jinhong Wu}
\author[1]{Tie Ji}
\author[1]{Yizhen Lao}

\affil[1]{Hunan University}
\affil[2]{Nanyang Technological University}

\teaser{
  \centering
  \includegraphics[width=0.99\textwidth]{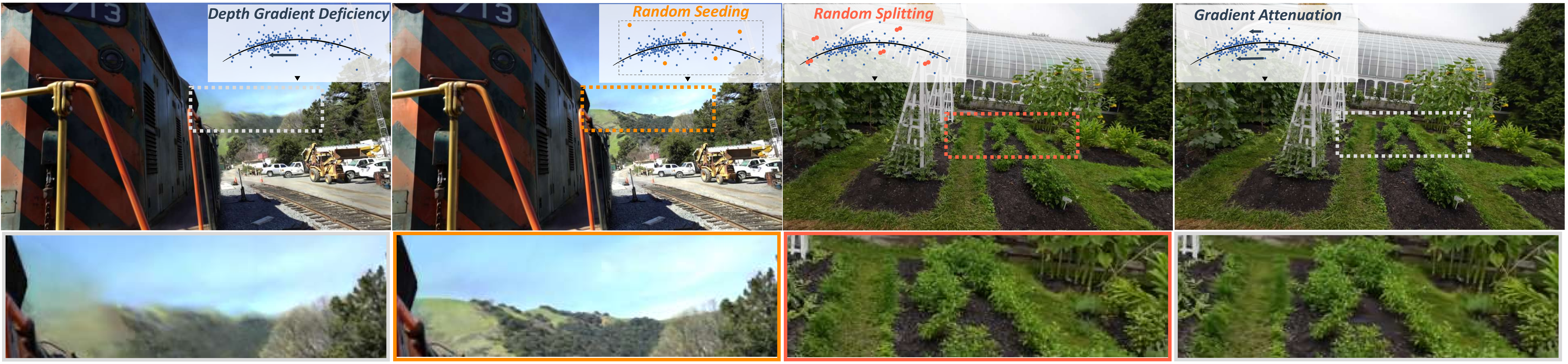}
  \vspace{-5pt}
  \captionof{figure}
  {
    \textbf{Escaping the Blur Trap via Stochastic Exploration.} 
    We identify the \textit{Blur Trap} as a fundamental limitation intrinsic to 3DGS, rooted in an inherent exploitation-only optimization bias. 
    It manifests as two variants: the Far-Side Blur Trap~(Left) triggered by depth-gradient deficiency and the Near-Side Blur Trap~(Right) driven by 2D-gradient attenuation. 
    To break this deadlock, we introduce minimal exploration operators: \textit{Random Seeding} and \textit{Random Splitting}. These straightforward interventions bypass gradient-dependent optimization, consistently recovering structural details and proving that simple explicit exploration is highly effective for escaping the Blur Trap without complex modifications.
  }
  \label{fig:teaser}
}

\begin{abstract}
3D Gaussian Splatting (3DGS) employs Gaussian primitives for explicit scene representation, facilitating real-time, high-fidelity reconstruction and novel view synthesis of complex scenes. 
However, the explicit modeling inherent in 3DGS introduces a gradient bias during optimization, rendering its non-convex optimization process highly susceptible to convergence toward local suboptimal solutions. \textbf{\textit{This constitutes a fundamental limitation in 3DGS optimization, which we term the Blur Trap}}.  
To address this limitation, we integrate simple explicit exploration into the 3DGS optimization framework. 
First, through rigorous mathematical analysis of the 3DGS optimization formulation, we identify the underlying optimization bias responsible for the Blur Trap and categorize it into two distinct subtypes: the Far-Side Blur Trap and the Near-Side Blur Trap. 
Subsequently, we propose two highly straightforward exploration strategies (\textit{Random Seeding} and \textit{Random Splitting}) to mitigate the far-side and near-side blur traps, respectively.  
Experimental validation demonstrates that the incorporation of these exploration operators effectively and complementarily overcome the Blur Trap, achieving high-quality rendering performance across multiple datasets. 
Project page: \url{https://chengbo-wang.github.io/ExploreGS/}.   
\end{abstract}

\begin{document}
\maketitle

\section{Introduction}
\label{sec_introduction}

3D Gaussian Splatting (3DGS)~\cite{kerbl20233d} has rapidly become a leading approach for novel view synthesis, offering real-time rendering together with photorealistic fidelity through an explicit representation built from anisotropic Gaussian primitives. Unlike implicit neural representations such as NeRF~\cite{mildenhall2021nerf}, 3DGS owes much of its efficiency to a physically grounded design choice: each Gaussian is updated by gradients derived directly from the camera projection model, with no intermediate neural network in the loop. This design has driven impressive progress on standard benchmarks. Yet a recurring failure pattern can be observed across 3DGS reconstructions. \textbf{\textit{Certain regions in the rendered scene remain visibly blurry no matter how many training views are provided.}} As shown in ~\Figure~\ref{fig:teaser}, this is not random noise. It appears consistently in two kinds of regions: distant content such as mountains and skylines, and occluded near-field regions such as the grass behind the rack. The persistence of these artifacts even under abundant supervision indicates that the cause is not insufficient data but something rooted in the optimization process itself. We refer to this systematic failure as the \textbf{Blur Trap}.

To understand why the Blur Trap forms, we trace the issue back to the gradient that drives 3DGS optimization. The 3D position update of each Gaussian is composed of three branches that correspond to the 2D projected position, the 2D covariance, and the spherical harmonic coefficients. Empirically, the 2D position component dominates the other two by two to three orders of magnitude across the entire training process, so \textbf{\textit{the 3D position update is effectively governed by 2D reprojection error alone}}. We then ask what this dominant component looks like geometrically. 
Through a careful derivation of the camera projection pipeline, we prove that \textbf{\textit{the 3D position update derived from the 2D gradient is strictly orthogonal to the viewing ray from the camera center to the Gaussian}}. The practical consequence is that the optimizer can only move primitives within the plane perpendicular to the viewing ray and never along the depth direction of the viewing ray. Compounding this, the same 2D gradient also serves as the criterion for adaptive densification, and the $\alpha$-blending pipeline systematically attenuates this signal for occluded Gaussians, since \textbf{\textit{later-ranked primitives in the depth-sorted sequence receive transmittance-weighted gradients that decay rapidly}}. The Blur Trap therefore decomposes into two subtypes. The \emph{Far-Side Blur Trap} arises in distant regions where there is no depth-directed optimization signal to drive Gaussians to their correct depth. The \emph{Near-Side Blur Trap} arises in occluded foreground regions where the densification trigger is suppressed before it can fire.

Existing efforts to improve 3DGS have largely circled around the Blur Trap without naming or resolving it. 
A first line of work refines the densification criterion. AbsGS~\cite{ye2024absgs} addresses gradient direction conflicts among sub-pixel signals, Pixel-GS~\cite{zhang2024pixel} reweights gradients by pixel coverage, and similar variants~\cite{mallick2024taming, fang2024mini} adjust when and how clone or split operations fire. These methods all consume the same input: a gradient signal that already carries the two biases described above. Adjusting the threshold or the weighting does not change the fact that the signal itself lacks depth information and is attenuated by $\alpha$-blending. 
A second line of work, exemplified by HoGS~\cite{liu2025hogs}, reparameterizes Gaussian positions in homogeneous coordinates to better express far-field content. This changes how distant primitives are represented but does not introduce any new optimization signal in the depth direction, so the underlying gradient deficiency persists. 
A third line of work, most notably 3DGS-MCMC~\cite{kheradmand20243d}, injects Langevin-style noise into Gaussian positions and reinterprets the entire pipeline as MCMC sampling. This is the closest in spirit to our perspective. However, the noise there is applied uniformly across the scene and is not tied to a specific diagnosis of why 3DGS fails. As a result, the method treats all primitives with the same exploratory mechanism, even though, as we show, the failure has two distinct geometric origins that call for two different remedies.

These observations suggest a simple reframing. 3DGS optimization is fundamentally a high-dimensional non-convex problem, and the broader optimization literature has long recognized that escaping local optima in such landscapes requires some form of exploration, whether through stochastic gradient noise~\cite{welling2011bayesian, jin2017escape}, perturbed gradient descent~\cite{ge2015escaping}, or simulated annealing. Standard neural network training enjoys several natural sources of such exploration, including the noise inherent in mini-batch stochastic gradients and implicit regularization from dropout or weight decay. 3DGS enjoys almost none of these. Training proceeds with a minibatch size of one, the update rule contains no implicit regularization, and as established above, the physically derived gradient is itself structurally biased in ways that prevent the optimizer from probing either depth or occluded regions. From this vantage point, the Blur Trap is not a surprise but the predictable outcome of running a deterministic, gradient-dominated optimizer on a landscape whose dominant signal cannot point in the directions that matter. The natural remedy is to put exploration back in, and to design it so that it directly compensates for what the gradient cannot supply.

Guided by this principle, we introduce two minimal exploration operators, each tied to one subtype of the Blur Trap. \textbf{Random Seeding} addresses the Far-Side Blur Trap by periodically injecting candidate Gaussians at randomly sampled 3D positions across the scene. This bypasses the orthogonality constraint, since new primitives can land at depths that no existing Gaussian could have reached through gradient updates alone. 
Seeds that fall in geometrically invalid regions are removed by standard opacity pruning, and seeds that land in plausible regions are picked up by the regular optimization loop and refined. \textbf{Random Splitting} addresses the Near-Side Blur Trap by periodically splitting a random subset of Gaussians regardless of their accumulated 2D gradient, thereby bypassing the densification gate that $\alpha$-blending suppresses. Both operators are deliberately simple. We choose this minimal design because our goal is to verify that explicit exploration, on its own, is what 3DGS optimization is missing. A more sophisticated exploration strategy is certainly possible, but the burden of prove should first rest on the principle itself rather than on engineering. 
Experiments on Mip-NeRF~360~\cite{barron2022mip}, Tanks~\&~Temples~\cite{knapitsch2017tanks}, Deep Blending~\cite{hedman2018deep}, OMMO~\cite{lu2023large}, and DL3DV~\cite{ling2024dl3dv} confirm that each operator resolves its targeted form of blur, that the two operators are complementary, and that together they consistently improve rendering fidelity across all datasets with only marginal computational overhead.

\vspace{-0.5\baselineskip}
\contribution{
Contributions of this paper can be summarized as:
\vspace{-0.2\baselineskip}
\begin{enumerate}[leftmargin=*, itemsep=0em]
    \item We identify and formalize the \textbf{Blur Trap}, a fundamental optimization limitation of 3DGS that arises from gradient biases inherent to its physically derived backpropagation. We further categorize the Blur Trap into two subtypes, the Far-Side Blur Trap and the Near-Side Blur Trap, each rooted in a different facet of the gradient bias.
    \item We provide a rigorous theoretical analysis of the Blur Trap. In particular, we prove that the 3D position update derived from the 2D gradient is orthogonal to the viewing ray, which directly explains the absence of depth-directed optimization, and we characterize how $\alpha$-blending systematically attenuates the densification signal for occluded primitives.
    \item We propose two minimal exploration operators, \textbf{Random Seeding} and \textbf{Random Splitting}, targeting the Far-Side Blur Trap and Near-Side Blur Trap respectively. The minimal design is intentional, isolating the contribution of exploration itself from incidental engineering choices.
    \item Extensive experiments on five different datasets demonstrate that each operator resolves its targeted form of blur, and that the two operators are complementary, jointly establishing exploration as a missing and essential ingredient in 3DGS optimization.
\end{enumerate}
}
\vspace{-0.75\baselineskip}

\section{Related Work}
\label{Related Work}

\subsection{Exploitation-Driven Gaussian Splatting Optimization}

\textit{Adaptive Density Control in 3D Gaussian Splatting.}
Adaptive density control~(ADC) is the cornerstone of 3D Gaussian Splatting~(3DGS), dynamically adjusting primitive counts to balance geometric fidelity and computational efficiency. 
However, its strict reliance on view-space positional gradients introduces fundamental optimization bottlenecks, including signal dilution, directional conflicts, and premature stagnation.
A primary line of research targets these gradient computation flaws. AbsGS~\cite{ye2024absgs} identifies gradient collisions and proposes homodirectional view-space gradients to mitigate directional conflicts. GDAGS~\cite{zhou2026gradient} introduce a Gradient Coherence Ratio to measure directional consistency, enabling selective densification by distinguishing over-reconstructed from under-densified regions. 
Addressing scale bias, Pixel-GS~\cite{zhang2024pixel} propose pixel-aware gradients to normalize magnitude variations caused by differing pixel footprints. Similarly, ~\cite{rota2024revising} replace heuristic gradient thresholds with pixel-error-driven criteria and correct opacity biases during cloning.

\textit{Modeling Distant Content in Unbounded Scenes.}
HoGS~\cite{liu2025hogs} addressed this by introducing homogeneous coordinates for position and scale, enabling unified near-far representation. Structural approaches have also proven effective: Scaffold-GS~\cite{lu2024scaffold} established an anchor-based hierarchical representation with view-adaptive attribute decoding, naturally handling varying observation distances. 
Proxy-GS~\cite{gao2026proxy} further refined anchor placement by projecting anchors onto proxy mesh surfaces, preventing redundant growth behind occluders.

While these advances enhance densification robustness and extend 3DGS to unbounded domains via coordinate transformations or structural priors, they remain fundamentally bound to deterministic, gradient-driven exploitation. 
Stochastic alternatives that step outside this paradigm are comparatively rare, and we discuss them in the following section.

\subsection{Exploration in Machine Learning}

\textit{Exploration as a learning principle.}  
The role of exploration is a recurring theme across non-convex optimization and reinforcement learning. When a purely greedy update rule operates on a complex landscape, it tends to commit prematurely to whatever direction the local signal points toward, missing solutions hidden behind barriers or in under-visited regions. Classical optimization addresses this with simulated annealing~\cite{kirkpatrick1983optimization}, while modern stochastic optimization adds gradient noise through Stochastic Gradient Langevin Dynamics~(SGLD)~\cite{welling2011bayesian} or perturbed gradient descent, which is shown to escape saddle points with near-dimension-free guarantees~\cite{ge2015escaping,jin2017escape}. Reinforcement learning takes the same idea further, treating exploration as a first-class algorithmic problem rather than an auxiliary mechanism~\cite{sutton1998reinforcement}.

\textit{Two families of exploration mechanisms.}  
Exploration mechanisms in this broader literature divide into two complementary families. \emph{Noise-based} methods inject stochasticity directly into the update rule, ranging from $\epsilon$-greedy action selection to parameter-space perturbation that adds noise to network weights~\cite{plappert2017parameter,fortunato2018noisy}. \emph{Coverage-driven} methods instead steer exploration by explicit estimates of novelty or uncertainty, as in count-based bonuses~\cite{bellemare2016unifying}, prediction-error curiosity~\cite{pathak2017curiosity}, and random network distillation~\cite{burda2018exploration}. The two families are not interchangeable: the former refines locally around the current state, while the latter allocates effort toward regions the model has never reached~\cite{osband2019deep}. Which is more effective depends on the specific failure mode the learner faces.

\textit{Exploration in 3DGS optimization.}  
Despite this rich literature, exploration remains a niche thread in 3DGS, where mainstream training follows a deterministic densify-and-prune cycle from a Structure-from-Motion initialization~\cite{kerbl20233d}. 3DGS-MCMC~\cite{kheradmand20243d} is the closest predecessor of our perspective: it reinterprets Gaussians as samples from a distribution and converts position updates into SGLD steps via Langevin noise. A contemporaneous work, Opt3DGS~\cite{huang2026opt3dgs}, makes the exploration--exploitation framing explicit by alternating an adaptive SGLD phase with a curvature-aware exploitation phase. Both apply a single, uniform perturbation across all Gaussians without distinguishing between distinct failure modes. In contrast, we trace the difficulty back to two specific gradient biases identified in \Section~\ref{ssec:blur_trap}, and design two minimal operators each tied to one bias: 
\textbf{Random Seeding} acts as a coverage-driven mechanism that introduces new primitives in depth intervals unreachable from existing Gaussians, while \textbf{Random Splitting} acts as a noise-based mechanism that bypasses the densification gate suppressed by $\alpha$-blending. 
Our work thereby treats exploration in 3DGS as a diagnosis-driven design problem rather than a uniform perturbation strategy.

\section{Understanding the Blur Trap in 3D Gaussian Splatting}
\label{sec:optimization_analysis}

In \Section~\ref{sec:optimization_analysis}, we present a systematic analysis of the optimization deficiencies in 3D Gaussian Splatting (3DGS) that lead to the Blur Trap. 
Specifically, \Section~\ref{ssec:greedy_exploitation} characterizes how the backpropagation in 3DGS induces an 2D gradient exploitation-dominant optimization regime. 
\Section~\ref{ssec:depth_gradient_densify_failure} then identifies the optimization bias of 3DGS.
Finally, \Section~\ref{ssec:blur_trap} analyses how these biases drives optimization into the Blur Trap.

\subsection{Screen-Space Gradient Dominance in Optimization}
\label{ssec:greedy_exploitation}

3D Gaussian Splatting (3DGS) reconstructs 3D scenes through a differentiable rendering framework. The forward pass builds a rendering pipeline that projects 3D Gaussian primitives onto the image plane via a \textbf{\textit{physical camera model}}, producing pixel colors through $\alpha$-blending. Crucially, backpropagation derives gradients through this same differentiable pipeline, enabling direct optimization of Gaussian attributes against the reconstruction loss. 
Among these attributes, the 3D position of each Gaussian primitive is paramount: it determines where the primitive projects in screen space, governs how it blends with neighboring Gaussians, and ultimately defines the geometric structure of the reconstructed scene~\cite{kotovenko2026edgs,lan20253dgs2}.

The 3D position gradient of a Gaussian, \(\mathbf{g}_{\mathrm{all}}\), decomposes into three components:
\begin{equation}
    \mathbf{g}_{\mathrm{all}} =
    \mathbf{g}_{\mathrm{2d}} + \mathbf{g}_{\mathrm{cov2d}} + \mathbf{g}_{\mathrm{sh}},
    \label{eq:g_all}
\end{equation}
where \(\mathbf{g}_{\mathrm{2d}}\) originates from the 2D screen-space projection, \(\mathbf{g}_{\mathrm{cov2d}}\) stems from the 2D covariance matrix governing the image-plane footprint, and \(\mathbf{g}_{\mathrm{sh}}\) derives from the spherical harmonic coefficients encoding view-dependent appearance. Conceptually, \(\mathbf{g}_{\mathrm{2d}}\) captures how the primitive should move to reduce reprojection error, while \(\mathbf{g}_{\mathrm{cov2d}}\) and \(\mathbf{g}_{\mathrm{sh}}\) convey complementary geometric and appearance signals.

\begin{figure}[!t]
    \centering
    \includegraphics[trim={0.cm 0 0.cm 0},clip,width=0.9\linewidth]{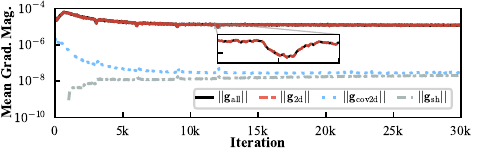}
    \vspace{-10pt}
    \caption{\textbf{Position gradient magnitude analysis.}
     Across the entire training process, $\|\mathbf{\textbf{g}}_{\text{2d}}\|$ remains almost identical to $\|\mathbf{\textbf{g}}_{\text{all}}\|$, while $\|\mathbf{\textbf{g}}_{\text{cov2d}}\|$ and $\|\mathbf{\textbf{g}}_{\text{sh}}\|$ are two to three orders of magnitude smaller, \textbf{demonstrating that the 2D gradient component dominates 3D position gradient update}.
    }
    \label{fig:pos_grad_mag}
\end{figure}

These three components differ drastically in magnitude. As shown in~\Figure~\ref{fig:pos_grad_mag}, \(\mathbf{g}_{\mathrm{2d}}\) is consistently orders of magnitude larger than \(\mathbf{g}_{\mathrm{cov2d}}\) and \(\mathbf{g}_{\mathrm{sh}}\). This disparity is structural: the 2D position gradient directly captures reprojection sensitivity to spatial displacements, whereas the covariance and SH gradients derive from less direct transformations. Consequently, \(\mathbf{g}_{\mathrm{2d}}\) dominates parameter updates, reducing the three-component optimization to greedily minimizing 2D reprojection error.

To verify this dominance, we ablate individual gradient components during Gaussian primitive updates. Models retaining \(\mathbf{g}_{\mathrm{2d}}\) maintain reconstruction performance comparable to standard 3DGS, confirming its sufficient role in driving optimization. Excluding \(\mathbf{g}_{\mathrm{2d}}\), however, leads to rapid quality degradation, while removing \(\mathbf{g}_{\mathrm{cov2d}}\) or \(\mathbf{g}_{\mathrm{sh}}\) individually produces negligible effect (Table~\ref{tab:pos_grad_items}). These results confirm that the optimization is effectively driven by screen-space positional signals alone, with the remaining components playing a marginal role.
In summary, although the 3D position update formally combines three gradient contributions, \textbf{their severe magnitude disparity degenerates the optimization into a greedy, 2D screen-space gradient exploitation process.} The optimizer minimizes reprojection errors through \(\mathbf{g}_{\mathrm{2d}}\) while effectively ignoring the complementary geometric and appearance signals from \(\mathbf{g}_{\mathrm{cov2d}}\) and \(\mathbf{g}_{\mathrm{sh}}\). As a result, Gaussian primitives rapidly converge to configurations that locally minimize pixel error. This is a classic exploitation behavior that prioritizes immediate screen-space fidelity at the expense of global structural integrity. \textbf{\textit{This analytical insight naturally leads to the central question: does such a greedy 2D optimization strategy introduce systematic reconstruction failures in 3DGS?}}

\begin{table}[!t]
  \centering
  \resizebox{1.0\linewidth}{!}{ 
  \begin{tabular}{@{}ccccccccccccc@{}}
    \toprule
        \multicolumn{3}{c}{\textbf{Pos. Grad. Item}} &
        \multicolumn{3}{c}{\textbf{Mip-NeRF360}~\cite{barron2022mip}} & 
        \multicolumn{3}{c}{\textbf{Tanks~\&Temples}~\cite{knapitsch2017tanks}} & 
        \multicolumn{3}{c}{\textbf{Deep Blending}~\cite{hedman2018deep}} \\
        \cmidrule(r){1-3} \cmidrule(l){4-6} \cmidrule(l){7-9}  \cmidrule(l){10-12} 
        \textbf{Cov.} & \textbf{2D} & \textbf{SH} & 
        \textbf{PSNR}\(\uparrow \) & \textbf{SSIM}\(\uparrow \) & \textbf{LPIPS}\(\downarrow \) & 
        \textbf{PSNR}\(\uparrow \) & \textbf{SSIM}\(\uparrow \) & \textbf{LPIPS}\(\downarrow \) & 
        \textbf{PSNR}\(\uparrow \) & \textbf{SSIM}\(\uparrow \) & \textbf{LPIPS}\(\downarrow \) 
        \\ 
        \midrule
        \checkmark & \betterV{\checkmark} & \checkmark & \betterV{27.52} & \betterV{0.816} & \betterV{0.215} & 
        \betterV{23.73} & \betterV{0.853} & \betterV{0.169} & \betterV{29.80} & \betterV{0.907} & \betterV{0.238}  \\ 
        ~ & \betterV{\checkmark} & \checkmark & \betterV{27.52} & \betterV{0.816} & \betterV{0.215} & 
        \betterV{23.77} & \betterV{0.853} & \betterV{0.169} & \betterV{29.85} & \betterV{0.907} & \betterV{0.238}  \\ 
        \checkmark & ~ & \checkmark & 24.27 & 0.661 & 0.386 & 
        21.59 & 0.729 & 0.333 & 24.92 & 0.825 & 0.376  \\ 
        \checkmark & \betterV{\checkmark} & ~ & \betterV{27.53} & \betterV{0.816} & \betterV{0.215} & 
        \betterV{23.73} & \betterV{0.853} & \betterV{0.169} & \betterV{29.78} & \betterV{0.907} & \betterV{0.239}  \\ 
        \checkmark & ~ & ~ & 24.18 & 0.658 & 0.388 & 21.54 & 0.724 & 0.336 & 24.88 & 0.826 & 0.376  \\ 
        ~ & \betterV{\checkmark} & ~ & \betterV{27.54} & \betterV{0.816} & \betterV{0.215} & 
        \betterV{23.84} & \betterV{0.853} & \betterV{0.170} & \betterV{29.80} & \betterV{0.907} & \betterV{0.238}  \\ 
        ~ & ~ & \checkmark & 25.23 & 0.706 & 0.340 & 22.45 & 0.771 & 0.285 & 28.12 & 0.879 & 0.301 \\ 
    \bottomrule
  \end{tabular}
  }
  \caption{
    \textbf{Ablation study on 3D position gradient items.} 
    We evaluate the impact of different gradient items on rendering quality. 
    2D position gradient branch is a key branch for 3DGS optimization.
    Pos. Grad. Item denote Position Gradient Item.
    }
  \label{tab:pos_grad_items}
\end{table}

\subsection{Depth Gradient Deficiency and Blending-Induced Densification Failure}
\label{ssec:depth_gradient_densify_failure}

The 2D position gradient in 3DGS fulfills a dual role: it drives 3D positional updates and serves as the core criterion for adaptive density control~(ADC)~\cite{kerbl20233d}. 
Consequently, this single optimization branch exerts disproportionate control over the entire reconstruction pipeline. 
We systematically analyze the inherent pitfalls of greedily relying on the 2D gradient for both position optimization and densification. Our analysis establish a mechanistic basis for understanding why standard 3DGS frequently stagnates in the \textit{Blur Trap}.
We demonstrate that this exploitation-heavy strategy inherently lacks depth-directed supervision. When coupled with the \(\alpha \)-blending pipeline, the periodically accumulated 2D gradients used for densification are systematically diluted. 
\textit{The detailed derivation is provided in the \Appendix~\ref{ssec:appendix_A} and ~\ref{ssec:appendix_B}.}

\subsubsection{Depth Gradient Deficiency}

\begin{figure}[!t]
    \centering
    \includegraphics[trim={0.cm 0 0.cm 0},clip,width=0.16\linewidth]{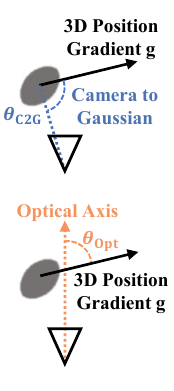}
    \includegraphics[trim={0.cm 0 0.cm 0},clip,width=0.82\linewidth]{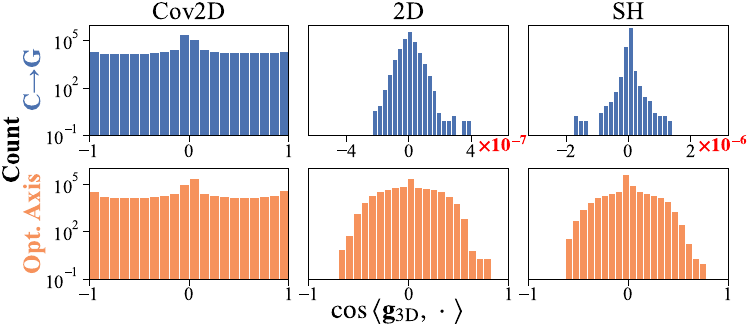}
    \caption{
    \textbf{Gradient Direction Analysis: Orthogonality and Depth-Signal Deficiency.} 
    Histograms of cosine similarity between the three 3D position gradient branches and the reference directions. 
    \textbf{Top:} Relative to the viewing ray. 
    The 2D position and SH branches are orthogonal to the viewing ray, confirming that screen-space reprojection errors cannot produce depth-directed updates. 
    \textbf{Bottom:} Relative to the optical axis. 
    Limited to field-of-view, optimization directions are mostly distributed between 60\(^{\circ}\) and 120\(^{\circ}\) relative to the optical axis, indicating a severe lack of depth-directed signals along the optical axis. 
    The Cov2D branch exhibits more diverse directional distribution, including depth-aligned components.
    However, its gradient magnitude is orders of magnitude smaller than that of the 2D position branch (\Section~\ref{ssec:greedy_exploitation}), rendering it ineffective for depth optimization. 
    This directional bias, combined with the magnitude imbalance, systematically denies 3DGS effective depth-directed gradient signals.
    }
    \label{fig:grad_direction}
\end{figure}

We prove that \textbf{the 3D positional update direction of any Gaussian primitive remains strictly orthogonal to the viewing ray from the camera center to the primitive}. This geometric constraint is formalized as:
\begin{equation}
\begin{aligned}
    \frac{\partial L^{2D}}{\partial \mathbf{P}_{3D}} (\mathbf{P}_{3D} - \mathbf{P}_{\mathrm{cam}})
    &= 0,
\end{aligned}
\label{eq:dL_d2D_d3D_0}
\end{equation}
where $\frac{\partial L^{2D}}{\partial \mathbf{P}_{3D}}$ denotes the 3D position gradient component derived exclusively from the 2D position gradient branch, $\mathbf{P}_{3D}$ represents the Gaussian's 3D position, and $\mathbf{P}_{\mathrm{cam}}$ indicates the camera optical center.

Translating a Gaussian along the viewing ray doesn't induce any change in its projected 2D screen position. Therefore, to minimize the loss from projection, the optimizer consistently directs 3D positional updates orthogonal to this ray. 
Given typically narrow camera fields of view and centrally framed subjects, viewing rays align closely with the optical axis. Consequently, the orthogonal update dirsections lie nearly parallel to the image plane. This geometric configuration inherently starves the optimization process of depth-directed signals.

\begin{figure}[!t]
    \centering
    \includegraphics[trim={0.cm 0 0.cm 0},clip,width=0.99\linewidth]{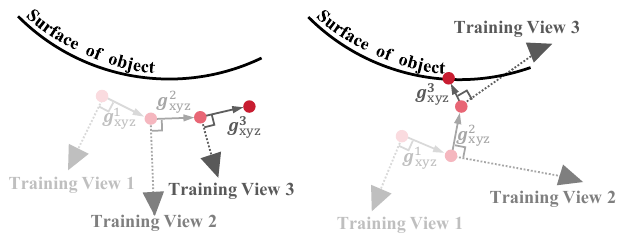}
    \caption{
    \textbf{Geometric Regularization via Multi-View Consistency under Gradient Orthogonality.} 
    3DGS optimizes Gaussian positions along directions strictly orthogonal to the viewing ray. 
    \textbf{Left:} Rich multi-view observations provide intersecting constraints that implicitly regularize primitives onto the true object surface. 
    \textbf{Right:} Under insufficient view diversity, this geometric regularization fails. 3DGS compensates by proliferating redundant primitives to fit limited view directions, leading to severe overfitting and structural artifacts. 
    Note that this is a simplified 2D illustration; in full 3D space, the absence of supervisory views from any single direction leads to incomplete geometric optimization.
    }
    \label{fig:grad_3d}
\end{figure}

We empirically validate this orthogonality and the resulting depth-gradient deficiency. As illustrated in~\Figure~\ref{fig:grad_direction}, we extract the 3D positional update vectors derived from three distinct gradient branches. We then compute their cosine similarities relative to both the viewing ray and the camera optical axis. The results confirm that updates driven by the 2D position gradient remain orthogonal to the Camera-to-Gaussian ray. Moreover, their cosine values against the optical axis cluster tightly around zero.
As summarized in~\Table~\ref{tab:grad_all_items}, the angular and magnitude characteristics confirm that the 3D position update is dominated by the 2D position gradient branch and remains orthogonal to the viewing ray. Consequently, optimization signals along the depth axis are substantially weaker than those within the image plane, and effective 3DGS reconstruction fundamentally depends on diverse camera trajectories. As shown in~\Figure~\ref{fig:grad_3d}, multi-view observations supply intersecting constraints that collectively resolve depth ambiguities.

\begin{table}[!t]
  \centering
  \resizebox{1.0\linewidth}{!}{ 
  \begin{tabular}{@{}cccl@{}}
    \toprule
        ~ & \textbf{Branch} & \textbf{Magnitude} & \textbf{Direction} \\ 
        \midrule
        ~ & \(\mathbf{\textbf{g}}_{\text{2d}}\) & \(\approx \|\mathbf{\textbf{g}}_{\text{all}}\|\) & Mathematically orthogonal to the viewing ray \\ 
        \(\mathbf{\textbf{g}}_{\text{all}}\)  & \(\mathbf{\textbf{g}}_{\text{sh}}\) & \(\ll \|\mathbf{\textbf{g}}_{\text{all}}\|\) & Experientially orthogonal to the viewing ray \\ 
        ~ & \(\mathbf{\textbf{g}}_{\text{cov2d}}\) & \(\ll \|\mathbf{\textbf{g}}_{\text{all}}\|\) & More random direction \\ 
    \bottomrule
  \end{tabular}
  }
  \caption{
    \textbf{Magnitude and direction of the three gradient branches.} 
    The 2D position gradient \(\mathbf{g}_{\mathrm{2d}}\) dominates the total position gradient in magnitude and is strictly orthogonal to the viewing ray, while the spherical harmonic and 2D covariance gradients contribute negligibly to the 3D position update.
    }
  \label{tab:grad_all_items}
\end{table}

In principle, capturing an object from all surrounding viewpoints provides sufficient directional supervision for full geometric optimization. In practice, however, acquiring multi-view coverage of every scene element is often prohibitively expensive (e.g., distant mountains in background) or physically impossible (e.g., clouds in the sky). In these cases, although the training set includes multiple views, the corresponding camera-to-object ray directions cluster within a narrow angular range. This limited viewing configuration deprives distant Gaussian primitives of depth-directed optimization signals, leading to reconstruction degradation in background regions~\cite{liu2025hogs}.

In summary, 3DGS optimization functions as an inherently local, position-driven process. Although the framework employs adaptive densification, it fundamentally lacks optimization momentum along the viewing ray. This absence of depth-directed exploration systematically constrains the effectiveness of the densification pipeline.

\subsubsection{Blending-Induced Densification Failure}

\begin{figure}[!t]
    \centering
    \includegraphics[trim={0.cm 0 0.cm 0},clip,width=0.99\linewidth]{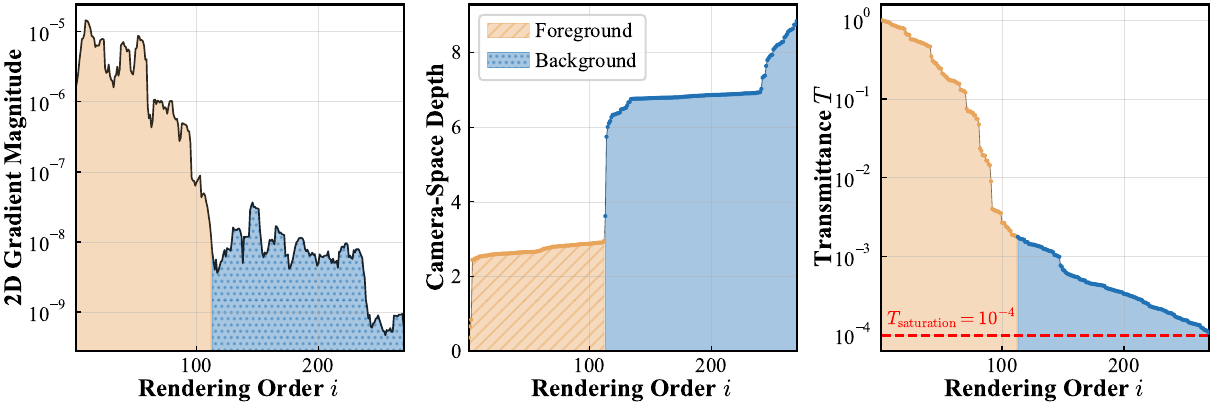}
    \caption{
    \textbf{Blending-Induced Gradient Attenuation and Color-Structure Inconsistency.} 
    The 2D position gradient magnitude, camera-space depth, and transmittance~$T$ of primitives compositing a single pixel, plotted against the rendering order $i$. 
    \textbf{Left:} \myorangetext{Foreground primitives}~($i \leq 112$) maintain strong gradient signals, while \mybluetext{background primitives}~($i \textgreater 112$) exhibit drastically attenuated magnitudes due to cumulative transmittance decay. 
    \textbf{Middle:} Foreground and background primitives occupy distinct depth intervals. Depth distribution reveals that although they jointly contribute to the same pixel color, they represent spatially disjoint geometric structures. 
    \textbf{Right:} Although foreground primitives significantly reduce accumulated transmittance $T$, $\alpha$-blending continues compositing background Gaussians until \(T < T_{\text{saturation}}\) (\(0.0001\)). This forces background primitives to density and optimize position with severely attenuated gradients.
    }
    \label{fig:blending_order}
\end{figure}

The 2D position gradient $\frac{\partial L}{\partial \mathbf{p}_{2D}}$ also acts as the screening metric for adaptive densification. Prior to each densification step, primitives exhibiting substantial gradient magnitudes on the image plane are flagged as spatially active and selected for densification. 
However, the \(\alpha \)-blending pipeline introduces a systematic gradient bias. \textbf{During backpropagation, \(\alpha \)-blending suppresses densification by delivering severely weakened 2D position signals to occluded regions}.

\begin{equation}
    C_p=\sum_{i=1}^N c_i \alpha_i T_i,\quad T_i=\prod_{j=1}^{i-1}\left(1-\alpha_j\right)
    \label{eq:blending}
\end{equation}

As formalized in \Equation~\ref{eq:blending}, where \(i\) denote the rendering order of a Gaussian within the \(\alpha \)-blending sequence. 
The primitives rendered later in the depth-sorted sequence contribute minimally to the final pixel color due to accumulated foreground opacity~\cite{zhu2026seele}. Consequently, the gradient backpropagated to these later-ranked Gaussians is consistently attenuated compared to that received by foreground primitives~\cite{wang2025faster}. The transmittance \(T_i\) effectively weights the contribution of primitive \(i\) to both the rendered output and the loss computation. \textbf{Since \(T_i\) decays rapidly as \(i\) increases, the 2D position optimization signal for occluded Gaussians is progressively suppressed.}

Rendering a single pixel typically requires compositing hundreds of Gaussian primitives via \(\alpha \)-blending~\cite{liu2026speeding}, governed by a predefined opacity termination criterion. 
However, these primitives that co-render the same pixel do not necessarily correspond to a single coherent 3D surface~\cite{gao2026proxy}, which means that the rendered pixels are inconsistent with the actual 3D structure~\cite{zhang2026rade,chen2024pgsr}.
Consequently, occluded background Gaussians share the same pixel as foreground primitives. 
Due to the multiplicative transmittance term in the \(\alpha \)-blending pipeline, gradient signals backpropagated to these occluded primitives are severely attenuated relative to foreground updates. 

We empirically validate this disparity by profiling the primitives contributing to a representative central pixel. Specifically, we record their 2D position gradient magnitudes and camera-space depths. As illustrated in \Figure~\ref{fig:blending_order}, primitives rendered earlier in the front-to-back sequence exhibit substantially larger gradient magnitudes than later-ranked ones. The corresponding depth visualization further reveals that these contributing Gaussians span three distinct depth intervals. 
Crucially, adaptive densification in 3DGS relies on periodically averaged 2D position gradients. For occluded primitives, this periodic statistic is systematically diluted by consistently weak signals across multiple views. As a result, the accumulated gradient fails to reach the splitting threshold, directly causing densification failure in background regions.

\subsection{Formation of the Blur Trap}
\label{ssec:blur_trap}

Collectively, screen-space gradient dominance~(\Section~\ref{ssec:greedy_exploitation}), depth gradient deficiency, and blending-induced densification failure~(\Section~\ref{ssec:depth_gradient_densify_failure}) establish an \textbf{exploitation-only optimization dynamic}. 
This dynamic fundamentally shapes the optimization landscape of 3DGS, consistently driving the system into persistent local optima. 

\begin{figure}[!t]
    \centering
    \includegraphics[trim={0.cm 0 0.cm 0},clip,width=0.99\linewidth]{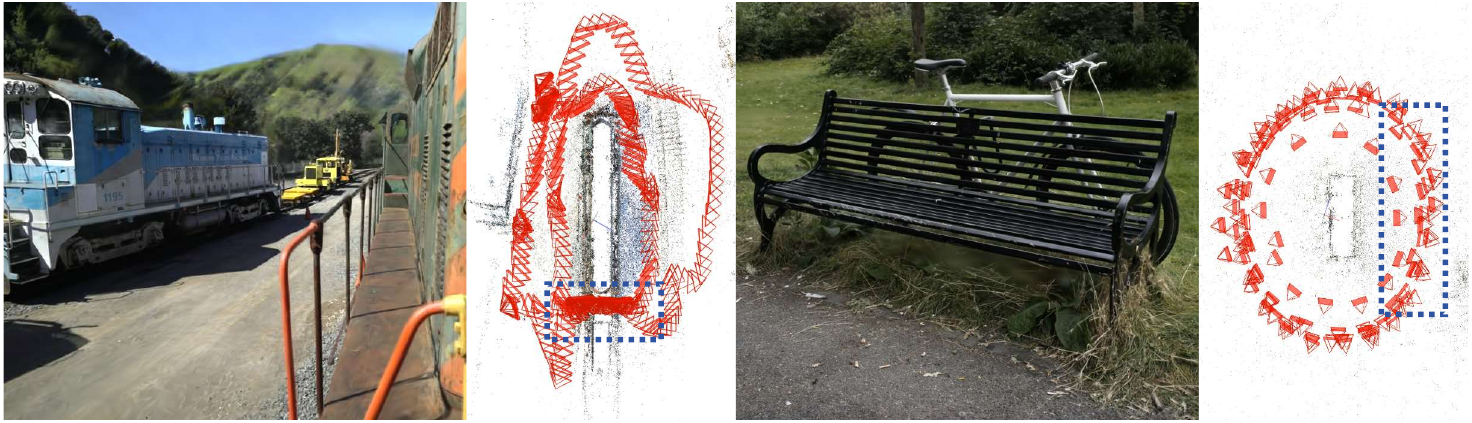}
    \caption{
    \textbf{Blur Trap under Sufficient Supervision.} 
    \textbf{Far-Side Blur Trap (Left):} 
    Distant regions (mountains) exhibit persistent structural smoothing even with abundant training views covering these areas. 
    \textbf{Near-Side Blur Trap (Right):} Occluded near-field areas (beneath the bench) lack structural detail despite adequate view coverage. 
    The dashed boxes in top-view camera pose distributions denote dense supervision views for the blurred areas. 
    }
    \label{fig:blur_trap}
\end{figure}

We identify this systematic optimization stagnation as a fundamental limitation intrinsic to 3DGS, which we formally define as the \textbf{Blur Trap}.
As illustrated in \Figure~\ref{fig:blur_trap}, pronounced blurring artifacts persist in distant regions (e.g., mountains) and occluded areas (e.g., beneath chairs), despite abundant multi-view supervision in the training views. 
Based on their distinct formation mechanisms, we categorize the Blur Trap into two variants: the \textit{Far-Side Blur Trap} and the \textit{Near-Side Blur Trap}.

\subsubsection{Optimization Dynamics and Trap Formation.} 

\paragraph{Far-Side Blur Trap.} 
The 3D position optimization in 3DGS inherently lacks depth-directed supervision in camera space. Consequently, Gaussian primitives fail to converge to their correct spatial locations along the viewing ray. This chronic absence of depth optimization signals directly manifests as the Far-Side Blur Trap.

\paragraph{Near-Side Blur Trap.} 
Although foreground primitives typically occupy more pixels and benefit from broader multi-view coverage, they frequently act as background elements in specific views. 
Under occlusion, the \(\alpha \)-blending pipeline systematically attenuates their 2D position gradients. 
Consequently, the periodically averaged gradient magnitude consistently falls below the densification threshold. 
This splitting stagnation traps local geometry in suboptimal configurations, ultimately forming the Near-Side Blur Trap.

\subsubsection{Motivation for Explicit Exploration.} 

The Blur Trap is not a transient optimization phase but an inherent consequence of the backpropagation pipeline. 
Deterministic gradient descent cannot escape this basin because the required depth-probing signals \textit{\textbf{are mathematically suppressed by the rendering formulation itself}}. 
Escaping this stagnation necessitates explicit perturbations.

\section{Methodology}
\label{sec:method}

In \Section~\ref{ssec:design_principles}, we first establish targeted exploration principles based on the formation mechanisms of the Blur Trap. 
Then, in \Section~\ref{ssec:random_seed} and \Section~\ref{ssec:random_split}, we introduce two minimal exploration operators designed to escape the Far-Side and Near-Side Blur Trap, respectively.

\subsection{Exploration Principles}
\label{ssec:design_principles}

Building on the analysis in \Section~\ref{sec:optimization_analysis}, we identify the root cause of the Blur Trap as the physics-based formulation of the differentiable rendering pipeline. 
Because both parameter updates and adaptive densification rely exclusively on backpropagated gradients, the optimization process inherits a systematic bias. 
Specifically, the overwhelming dominance of the 2D position gradient reduces 3D position optimization to a greedy, screen-space exploitation process. This gradient-dependent dynamic inherently neglects depth-ambiguous and occluded regions, ultimately trapping the reconstruction in suboptimal configurations.

To break this, our core design principle is to \textbf{bypass purely gradient-driven exploitation by injecting explicit exploration}. 
Rather than waiting for suppressed signals to recover, we proactively introduce perturbations that operate independently of the 2D gradient magnitude. 
Specifically, for the Far-Side Blur Trap, exploration must establish update pathways that propagate primitives into distant regions along the depth axis. For the Near-Side Blur Trap, it must proactively trigger densification for primitives whose 2D gradients consistently fall below the standard split threshold~\(\tau_\text{split}\).

\textit{\textbf{To rigorously validate the intrinsic efficacy of exploration, we deliberately adopt the simplest possible operators, prioritizing conceptual simplicity over engineering refinements}}.
Accordingly, we introduce two minimal exploration operators: \textit{Random Seeding} and \textit{Random Splitting}, detailed in \Section~\ref{ssec:random_seed} and \Section~\ref{ssec:random_split}
, which directly target the \textit{Far-Side} and \textit{Near-Side Blur Trap}, respectively.

\begin{figure}[!t]
    \centering
    \includegraphics[trim={0.cm 0 0.cm 0},clip,width=0.99\linewidth]{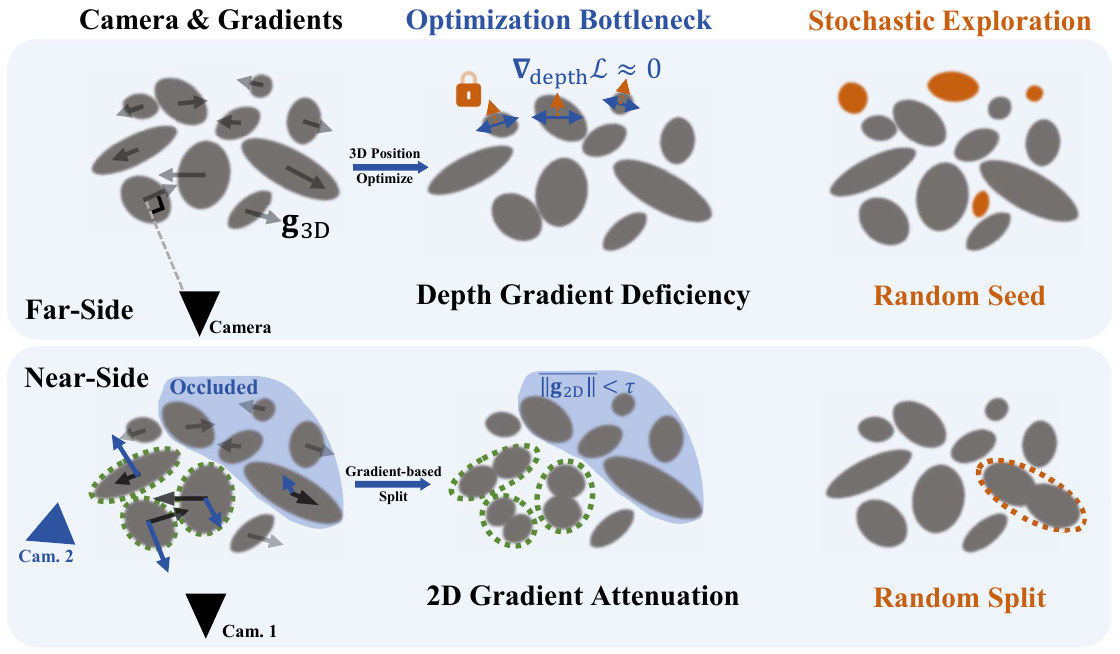}
    \caption{
    \textbf{Stochastic Exploration for Escaping the Blur Trap.} 
    \textbf{Far-Side~(Top):} 
    Physically-derived 3D position gradients $\mathbf{g}_{\text{3D}}$ are strictly orthogonal to the viewing direction, yielding zero update signals along the depth axis ($\nabla_{\text{depth}}\mathcal{L} \approx 0$) in distant zone.  
    \textit{Random Seed} injects \myorangetext{sparse Gaussian primitives} directly to probe unexplored depths. 
    \textbf{Near-Side~(Bottom):} 
    In \mybluetext{occluded regions}, $\alpha$-blending attenuation suppresses gradient magnitudes below the densification threshold ($\|\mathbf{g}_\text{2D}\| < \tau$). 
    Consequently, standard \mygreentext{gradient-based splitting} fails to allocate representational capacity to structurally important primitives. 
    \textit{Random Split} relaxes this deterministic mask via probabilistic splitting, \myorangetext{random splitting Gaussians} to restore detail. 
    }
    \label{fig:random_explore}
\end{figure}

\begin{table*}[!t]
  \centering
  \resizebox{1.0\linewidth}{!}{
  \begin{tabular}{@{}lcccccccccccccccc@{}}
    \toprule
        \multirow{2}{*}[-2pt]{\textbf{Method}} &
        \multicolumn{3}{c}{\textbf{Mip-NeRF360}~\cite{barron2022mip}} & 
        \multicolumn{3}{c}{\textbf{Tanks~\&~Temples}~\cite{knapitsch2017tanks}} & 
        \multicolumn{3}{c}{\textbf{Deep Blending}~\cite{hedman2018deep}}  & 
        \multicolumn{3}{c}{\textbf{OMMO}~\cite{lu2023large}}  & 
        \multicolumn{3}{c}{\textbf{DL3DV Benchmark}~\cite{ling2024dl3dv}} \\
        \cmidrule(r){2-4} \cmidrule(l){5-7} \cmidrule(l){8-10} \cmidrule(l){11-13} \cmidrule(l){14-16} 
         & 
        \textbf{PSNR}$\uparrow$ & \textbf{SSIM}$\uparrow$ & \textbf{LPIPS}$\downarrow$ & 
        \textbf{PSNR}$\uparrow$ & \textbf{SSIM}$\uparrow$ & \textbf{LPIPS}$\downarrow$ & 
        \textbf{PSNR}$\uparrow$ & \textbf{SSIM}$\uparrow$ & \textbf{LPIPS}$\downarrow$ & 
        \textbf{PSNR}$\uparrow$ & \textbf{SSIM}$\uparrow$ & \textbf{LPIPS}$\downarrow$ & 
        \textbf{PSNR}$\uparrow$ & \textbf{SSIM}$\uparrow$ & \textbf{LPIPS}$\downarrow$ 
        \\ 
        \midrule
        3DGS* & 
        27.52  & 0.816  & 0.215  & 
        23.73  & 0.853  & 0.169  & 
        \tidP{29.80}  & \tidP{0.907}  & \sndP{0.238}  & 
        30.49  & 0.920  & 0.142  & 
        27.16  & 0.860  & 0.167  \\ 
        HoGS*~(50K Iteration) & 
        27.55  & 0.818  & \tidP{0.201}  & 
        24.23  & 0.859  & \tidP{0.160}  & 
        29.21  & 0.898  & \tidP{0.244}  & 
        30.66  & 0.919  & 0.135  & 
        \tidP{28.16}  & \tidP{0.878}  & \tidP{0.140}  \\ 
        Seed Explore & 
        \tidP{27.68}  & \tidP{0.819}  & 0.214  &
        \sndP{24.30}  & \tidP{0.861}  & 0.163  & 
        29.78 & 0.906  & \fstP{0.237}  & 
        \tidP{30.85}  & \tidP{0.924}  & \tidP{0.131}  & 
        27.74  & 0.869  & 0.155  \\ 
        Split Explore & 
        \sndP{27.95}  & \fstP{0.829}  & \fstP{0.192}  & 
        \sndP{24.30}  & \sndP{0.868}  & \fstP{0.139}  & 
        \fstP{30.01}  & \fstP{0.908}  & 0.248  & 
        \fstP{31.29}  & \sndP{0.928}  & \sndP{0.121}  & 
        \fstP{28.47}  & \sndP{0.889}  & \sndP{0.125}  \\ 
        Seed~\&~Split Explore & 
        \fstP{27.96}  & \fstP{0.829}  & \sndP{0.195}  & 
        \fstP{24.37}  & \fstP{0.870}  & \fstP{0.139}  & 
        \sndP{29.98}  & \sndP{0.907}  & 0.249  & 
        \sndP{31.27}  & \fstP{0.930}  & \fstP{0.119}  & 
        \sndP{28.43}  & \fstP{0.891}  & \fstP{0.124} \\ 
    \bottomrule
  \end{tabular}
  }
  \caption{
    \textbf{Reconstruction Quality Across Different Datasets.} 
    \textit{Seed Exploration} consistently improves unbounded scenes (e.g., T\&T), while \textit{Split Exploration} excels in occluded regions (e.g., OMMO). 
    Their combination achieves optimal performance in most scenarios, validating simultaneous resolution of both Blur Trap variants. 
    (*All experiments rerun based on the same data.)
    }
  \label{tab:exp_results}
\end{table*}

\begin{table}[!t]
  \centering
  \resizebox{0.85\linewidth}{!}{ 
  \begin{tabular}{@{}lccccc@{}}
    \toprule
        \textbf{Method} & \textbf{Mip-360} & \textbf{T~\&~T} & 
        \textbf{DB} & \textbf{OMMO} & \textbf{DL3DV} \\
        \midrule
        3DGS & 2.72  & 1.57  & 2.48  & 1.78  & 1.14  \\ 
        HoGS~(50K) & 4.40  & 2.16  & 2.43  & 2.01  & 1.61  \\ 
        Seed Exp. & 2.62  & 1.51  & 2.33  & 1.68  & 1.11  \\ 
        Split Exp. & 2.58  & 2.13  & \textbf{0.86}  & 1.77  & 2.13  \\ 
        Seed~\&~Split Exp. & 2.53  & 2.11  & \textbf{0.79}  & 1.77  & 2.11 \\ 
    \bottomrule
  \end{tabular}
  }
  \caption{
    \textbf{Gaussian primitive allocation (in millions) across different models.} 
    \textit{Seed Exploration (Seed Exp.)} marginally reduces the primitive count compared to 3DGS. 
    On the Deep Blending~(DB) dataset, \textit{Split Exploration (Split Exp.)} drastically decreases Gaussian count. Conversely, for complex scenes like Tanks~\&~Temples (T\&T) and DL3DV, the strategy adaptively increases primitive allocation to resolve fine geometric details, resulting in a substantial fidelity gain (\Table~\ref{tab:exp_results}). 
    }
  \label{tab:exp_num_G}
\end{table}

\subsection{Random Seeding for Far-Side Exploration}
\label{ssec:random_seed}

Guided by the principle of establishing depth-directed pathways, the \textit{Random Seeding} operator directly stochastically samples a sparse set of seed Gaussians in 3D space and injects them into the existing primitive collection, as illustrated in the top row of~\Figure~\ref{fig:random_explore}. 
This intervention does not violate the viewing-gradient orthogonality inherent to 3DGS. By distributing seeds uniformly across the 3D space, the operator explicitly probes depth intervals that are inaccessible to gradient-based updates, thereby compensating for the lack of depth-directed optimization signals in the differentiable framework. 

The sparse injection schedule ensures minimal disruption to well-reconstructed regions, preserving the stability of the ongoing optimization. Seeds placed in invalid or empty areas fail to reduce the photometric loss and are automatically removed by the native opacity pruning mechanism. Conversely, seeds landing in geometrically plausible regions are guided by the standard 2D reprojection loss. They refine their positions within the plane orthogonal to the viewing ray, eventually triggering adaptive densification. This establishes a complementary optimization loop: \textbf{\textit{Random Seeding} supplies global depth exploration, while deterministic gradients drive local planar exploitation}. 

We characterize this operator as a global exploration mechanism. Its primary function is to overcome the fundamental limitation of 3DGS: the inability of deterministic, locally-focused position updates to explore uncharted depth dimensions.

\vspace{-10pt}
\subsection{Random Splitting for Near-Side Exploration}
\label{ssec:random_split}

As illustrated in the bottom row of~\Figure~\ref{fig:random_explore}, the \textit{Random Splitting} operator bypasses the gradient-dependent densification criterion by proactively targeting large-scale Gaussians across the entire scene for exploratory splitting. 
Specifically, we sample primitives based on their mean scale and split only a small subset ($N_{\text{split}}$) per iteration to preserve training stability. This strategy allows occluded regions to circumvent the suppression of 2D gradients and actively participate in the densification process. Once these regions are populated with finer primitives, they rapidly converge to optimal configurations under unoccluded training views. 

\textit{Random Splitting} directly addresses the fundamental limitation of 3DGS: in occluded areas, the contribution to foreground color is attenuated by $\alpha$-blending, which dilutes the 2D position signals. 
Consequently, these primitives fail to reach the statistical splitting threshold, preventing effective densification. \textbf{\textit{Random Splitting} restores this capability by decoupling structural refinement from gradient magnitude.}

\section{Experiments}

\label{sec:experiments}

\begin{figure*}[!t]
    \centering
    \includegraphics[trim={0.cm 0 0.cm 0},clip,width=0.99\linewidth]{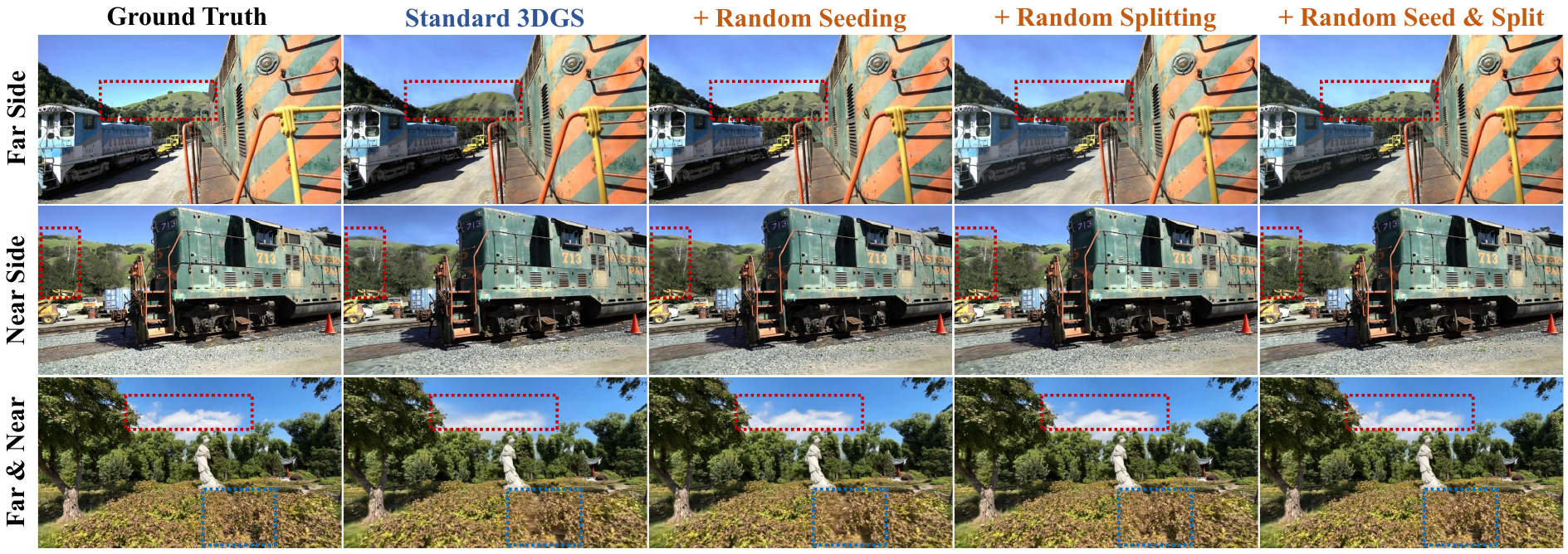}
    \caption{
    \textbf{Complementarity of Random Seeding and Random Splitting.}
    \textbf{Top: }
    Standard 3DGS produces persistent blur in far-side regions, while \textit{Random Seeding} effectively recovers these details by enabling depth-directed exploration.
    \textbf{Middle: }
    3DGS suffers from blur in occluded near-side regions, while \textit{Random Splitting} successfully restores structural sharpness by decoupling densification from gradient magnitude.
    \textbf{Bottom or fourth column: }
    \textit{Random Seeding} and \textit{Random Splitting} complement each other to escape both Blur Trap variants, recovering clear details in far-side backgrounds and occluded near-side regions.
    }
    \label{fig:far_near_com}
\end{figure*}

\subsection{Experimental Settings}

\subsubsection{Implementation Details}

We integrate these operators into standard 3DGS adhering to the principle of minimal intervention.

For \textit{Random Seeding}, the operator uniformly samples \(N_{\text{seed}}\) candidate positions at each densification iteration within the minimum bounding box of all existing Gaussians (\(N_{\text{seed}}=20\) by default).
For \textit{Random Splitting}, the operator randomly selects \(N_{\text{split}}\) large-scale Gaussians to split per iteration (\(N_{\text{split}}=20\) by default).
Although \(N_{\text{seed}}\) and \(N_{\text{split}}\) are negligible compared to the total primitive count (often in the millions), this sparse exploration proves highly effective in escaping the Blur Trap.

\subsubsection{Dataset and Metrics}

We comprehensively evaluate our method across multiple 3D reconstruction benchmarks, including Mip-NeRF 360~\cite{barron2022mip}, Tanks \& Temples~\cite{knapitsch2017tanks}, Deep Blending~\cite{hedman2018deep}, the large-scale outdoor dataset OMMO~\cite{lu2023large}, and the DL3DV Benchmark~\cite{ling2024dl3dv}. 
We randomly select a subset of scenes from the DL3DV Benchmark for detailed evaluation with $\frac{1}{4}$ resolution images~\footnote{DL3DV Scene IDs: 1, 101, 104, 109, 11, 21, 24, 26, 34, 42, 69, 7, 8, 82, 90, 97}. 
It is important to note that the Blur Trap is not unique to these selected scenes; rather, it is a widespread phenomenon observed across unselected scenarios in all tested datasets. 
Reconstruction performance is quantified using PSNR, SSIM, and LPIPS. Additionally, the number of Gaussian primitives is monitored to assess computational efficiency.

\subsection{Experiments Results}

\begin{figure*}[!t]
    \centering
    \vspace{10pt}
    \includegraphics[trim={0.cm 0 0.cm 0},clip,width=0.99\linewidth]{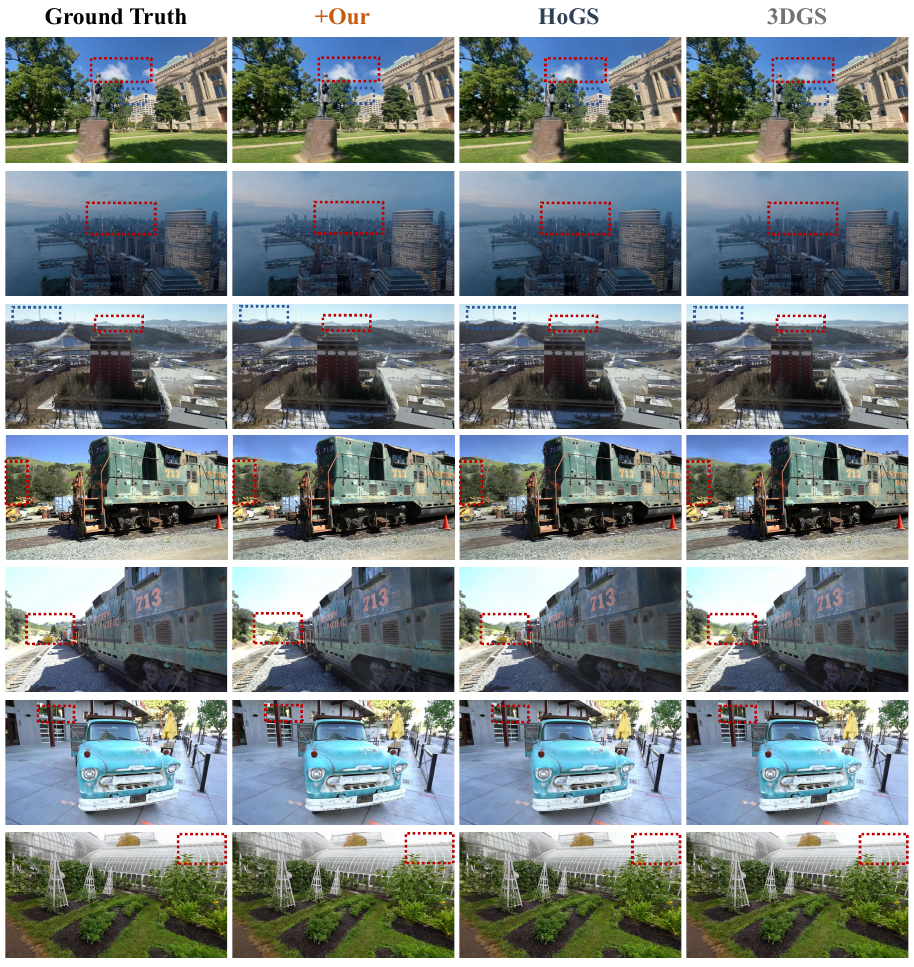}
    \vspace{10pt}
    \caption{
    \textbf{Visual comparison of escaping Far-Side Blur Trap for different models.}
    }
    \label{fig:fo_far}
\end{figure*}

\begin{figure*}[!t]
    \centering
    \vspace{10pt}
    \includegraphics[trim={0.cm 0 0.cm 0},clip,width=0.99\linewidth]{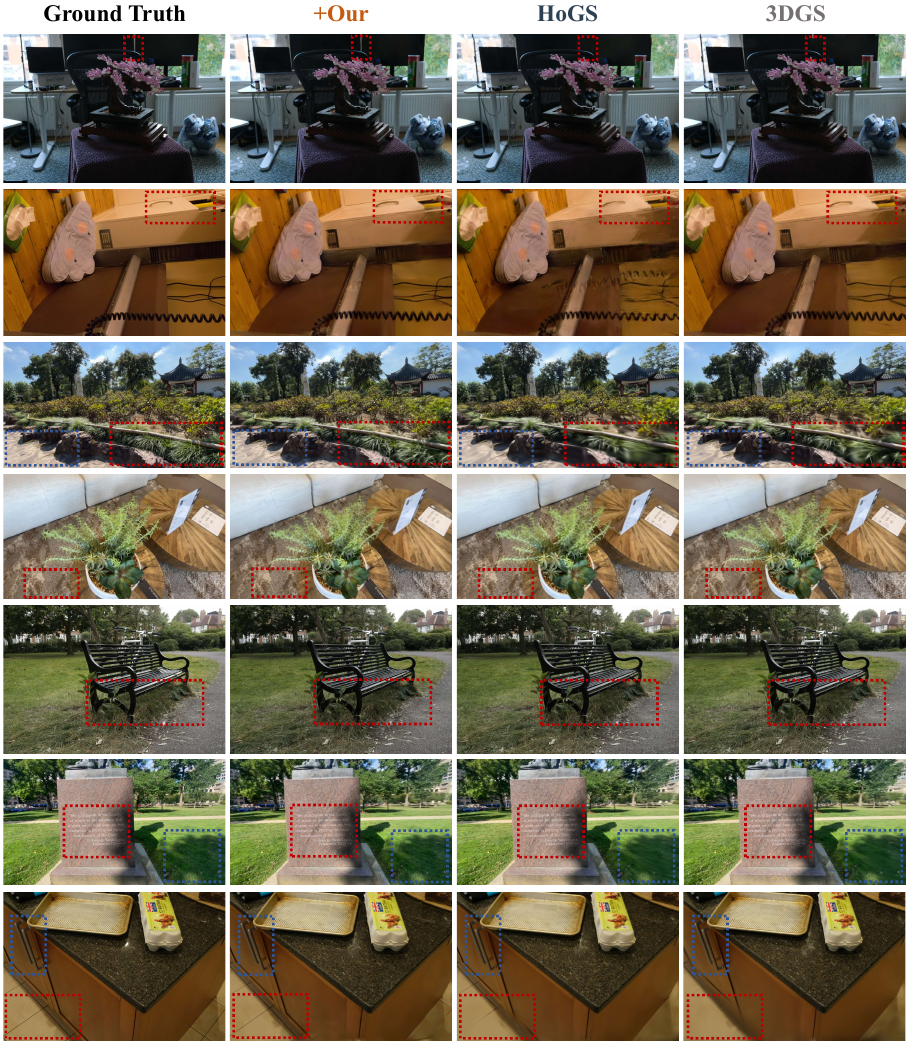}
    \caption{
    \textbf{Visual comparison of escaping Near-Side Blur Trap for different models.}
    }
    \label{fig:fo_near}
\end{figure*}

Quantitative results across multiple datasets confirm the efficacy of our minimal exploration operators. 
As shown in \Table~\ref{tab:exp_results}, both \textit{Seed Exploration} and \textit{Split Exploration} consistently improve reconstruction fidelity, with their combination achieving optimal performance across most benchmarks. This validates that the Far-Side and Near-Side Blur Traps can be simultaneously resolved through exploration.
The visual comparison is shown in \Figure~\ref{fig:fo_far} and  \Figure~\ref{fig:fo_near}.

The Deep Blending~(DB) dataset offers a critical insight. While \textit{Split Exploration} achieves the highest PSNR and SSIM, its LPIPS score is slightly suboptimal. This trade-off is directly explained by~\Table~\ref{tab:exp_num_G}, which reveals a 65\% reduction in Gaussian primitives compared to the 3DGS baseline, while maintaining superior geometric fidelity. This confirms that the operator prioritizes structural efficiency over redundant primitives, effectively resolving the Near-Side Blur Trap without introducing unnecessary complexity.

Crucially, \textit{Seed Exploration} requires only minimal modifications to the standard 3DGS pipeline. It slightly reduces the total primitive count while consistently improving PSNR, demonstrating that simple stochastic exploration is highly effective. 
Similarly, \textit{Split Exploration} reduces primitives in simpler regions (e.g., DB) while adaptively increasing them in geometrically complex scenes, a favorable trade-off that allocates computation where it most improves fidelity. This adaptive allocation reduces redundancy in simple areas while refining complex structures, enabling our operators to overcome the Near-Side Blur Trap.
The visual comparison in \Figure~\ref{fig:far_near_com} confirms complementarity: \textit{Random Seeding} recovers far-side background details that 3DGS leaves blurry, while \textit{Random Splitting} restores sharpness in occluded near-side regions. Their joint application (fourth column) simultaneously resolves both types of degradation, demonstrating that the two operators target distinct failure modes of the 3DGS optimization.

Together, these results confirm that minimal stochastic exploration is sufficient to escape the Blur Trap by disrupting its gradient-driven optimization dynamics. Our operators consistently improve fidelity while maintaining or reducing model complexity, establishing that \textbf{the Blur Trap arises from a fundamental optimization bias in 3DGS: heavy exploitation without explicit exploration}.

\subsection{4D Gaussian Splatting Blur Trap}
\label{ssec:4dgs}

\begin{figure}[!t]
    \centering
    \includegraphics[trim={0.cm 0 0.cm 0},clip,width=1.0\linewidth]{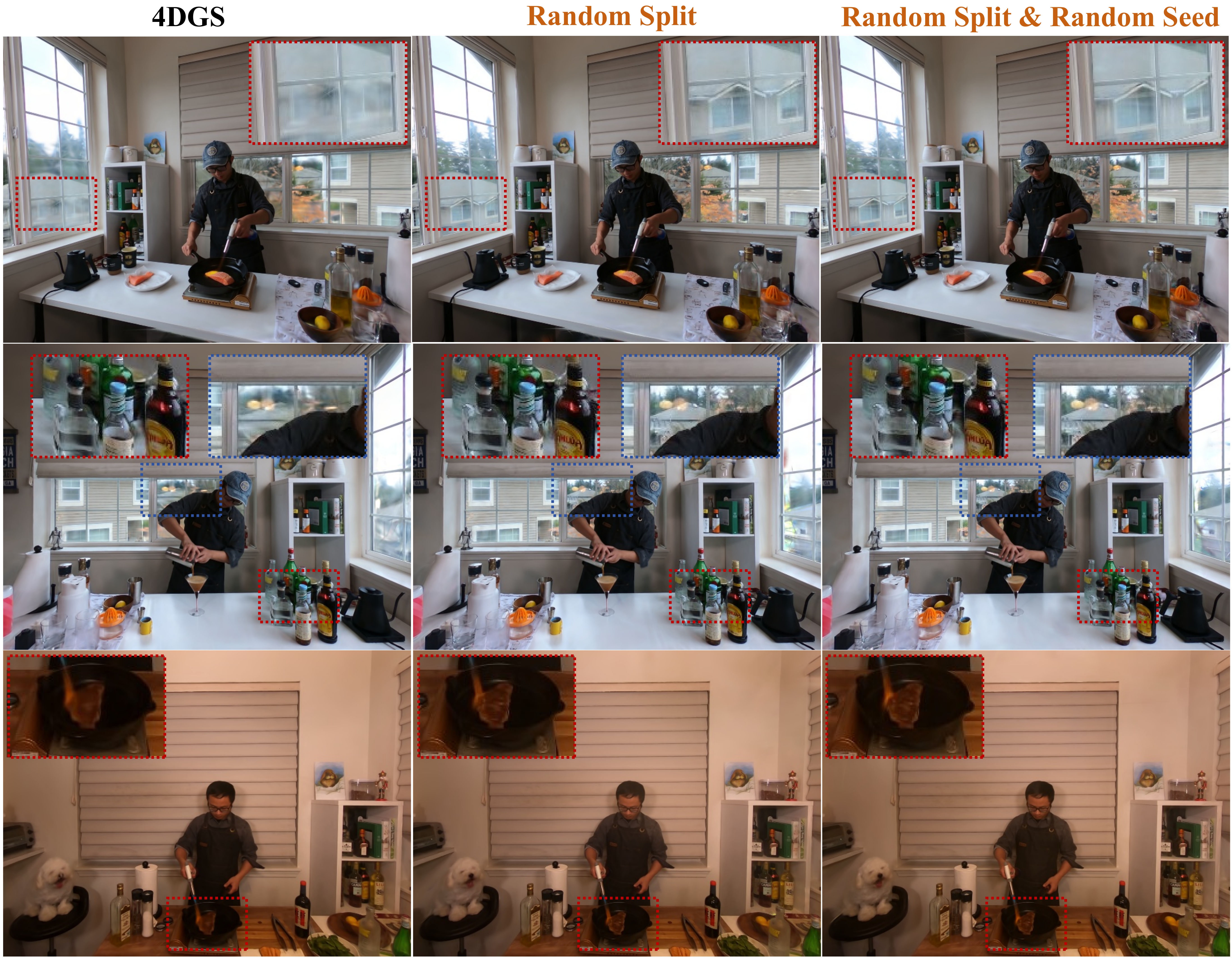}
    \caption{
    \textbf{Random Exploration in 4D Gaussian Splatting.}
    \textbf{Left: }
    4DGS inherits the Blur Trap, failing to densify occluded regions. Background structures (e.g., buildings behind windows) receive severely attenuated optimization signals that systematically suppress densification. Similarly, dynamic occluders such as flames hinder the refinement of food textures.
    \textbf{Middle and Right: }
    By integrating a highly sparse \textit{Random Splitting} ($N_{\text{split}}=5$), these stagnated primitives bypass gradient-dependent thresholds and undergo structural refinement. This minimal intervention yields substantial detail recovery across both static and dynamic occlusions, confirming that explicit exploration transfers to 4DGS pipelines.
    }
    \label{fig:4dgs_exp}
\end{figure}

\paragraph{4D Gaussian Splatting}
Due to the shared optimization pipeline, the Blur Trap inherently propagates from 3DGS to 4D Gaussian Splatting~(4DGS)~\cite{wu20244d}. We validate this limitation on a representative 4DGS framework with Neu3D~\cite{li2022neural} dataset. 
As illustrated in \Figure~\ref{fig:4dgs_exp}, despite extensive multi-view coverage, regions classically susceptible to occlusion remain severely blurred in the baseline reconstruction, such as distant buildings visible through windows, text on indoor bottles, and steak textures dynamically obscured by flames. These areas suffer from suppressed 2D positional gradients, which systematically inhibits adaptive densification.

We integrating \textit{Random Splitting} with a highly sparse splitting ($N_{\text{split}}=5$), and it successfully bypasses the Blur Trap to trigger effective densification. 
Occluded primitives actively participate in structural refinement and detail recovery. As quantified in \Table~\ref{tab:4dgs_exp}, these gains are most pronounced in perceptual fidelity (LPIPS). 
The fact that such a minimal intervention resolves severe blur confirms that standard splitting mechanisms, which strictly rely on 2D positional gradients, are fundamentally bottlenecked by signal attenuation. 
Explicit exploration, even at extremely low intensity, provides a more robust pathway for occluded-region refinement.

\begin{table}[!t]
  \centering
  \resizebox{0.99\linewidth}{!}{ 
  \begin{tabular}{@{}lccccccc@{}}
    \toprule
        \textbf{Method} & 
        \textbf{PSNR~\(\uparrow \)} & \textbf{SSIM~\(\uparrow \)} & \textbf{MS-SSIM~\(\uparrow \)} & 
        \textbf{D-SSIM~\(\downarrow \)} & \textbf{LPIPS-vgg~\(\downarrow \)} & \textbf{LPIPS-alex~\(\downarrow \)}  \\
    \midrule
    4DGS                & 
    30.575 & 0.9314 & 0.9659 & 0.0171 & 0.1507 & 0.0602 \\
    Split Exp.          & 
    \sndP{30.820} & \sndP{0.9369} & \sndP{0.9689} & \sndP{0.0156} & \sndP{0.1380} & \sndP{0.0488} \\
    Split \& Seed Exp.  & 
    \fstP{31.085} & \fstP{0.9379} & \fstP{0.9702} & \fstP{0.0149} & \fstP{0.1377} & \fstP{0.0483} \\
    \bottomrule
  \end{tabular}
  }
  \caption{
    \textbf{4DGS on the Neu3D Dataset.} 
    \textit{Random Splitting} delivers the primary fidelity gains. Occlusion-induced near side blur remains the optimization bottleneck in Neu3D~\cite{li2022neural}. These scenes are captured within a relatively bounded depth range, 
    which naturally limits the effectiveness of \textit{Random Seeding}. 
    }
  \label{tab:4dgs_exp}
\end{table}

\begin{figure}[!t]
    \centering
    \includegraphics[trim={0.cm 0 0.cm 0},clip,width=0.99\linewidth]{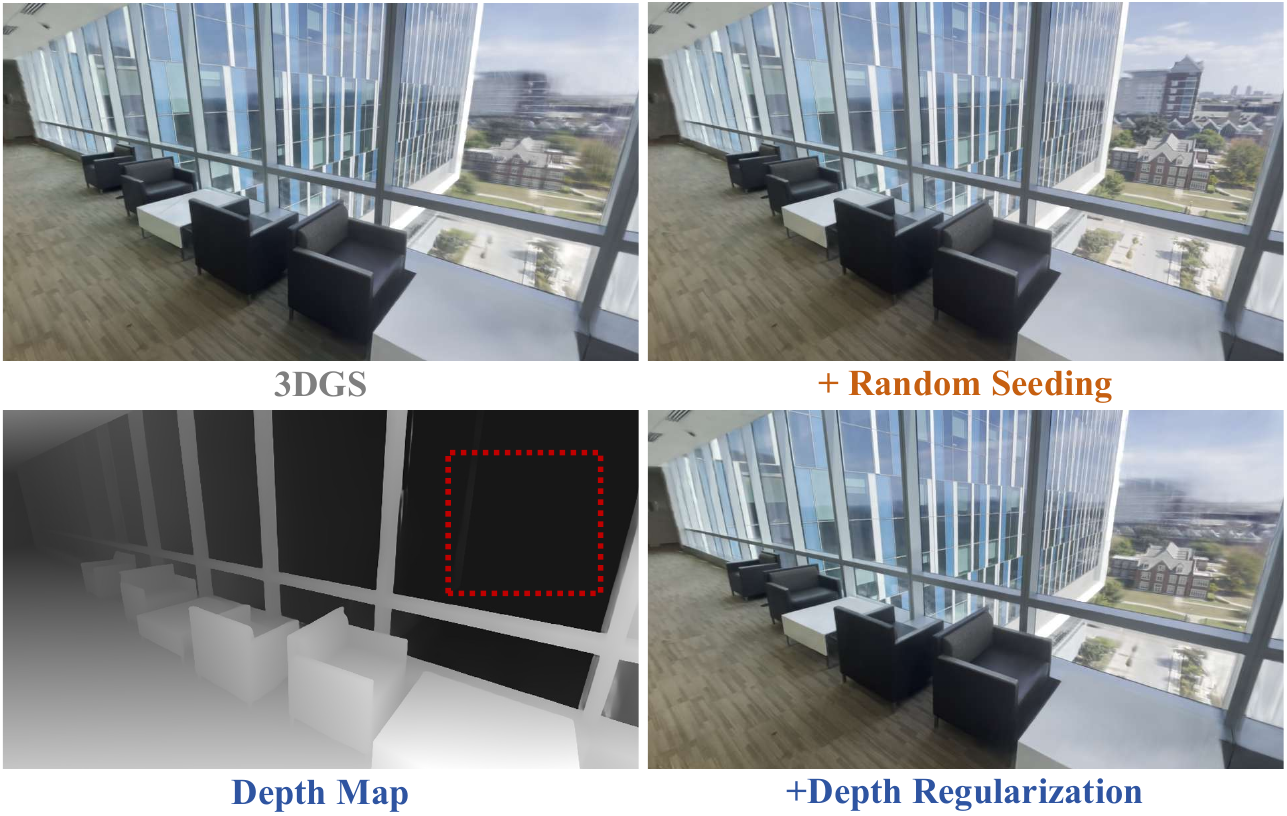}
    \caption{
    \textbf{Depth Supervision versus Random Seeding for Far-Side Recovery.} 
    \textbf{Top-Left}: Standard 3DGS suffers from the Far-Side Blur Trap due to absent depth-directed optimization signals. 
    \textbf{Buttom}: Incorporating pseudo-depth supervision partially alleviates distant blur, but its inherent inaccuracy restricts the effective range, leaving farther regions trapped. 
    \textbf{Top-Right}: In contrast, \textit{Random Seeding} explicitly probes uncharted depth intervals, enabling robust escape from the Far-Side Blur Trap across varying distances. 
    }
    \label{fig:depth_loss}
\end{figure}

\begin{table}[!t]
  \centering
  \resizebox{1.\linewidth}{!}{   
  \begin{tabular}{@{}clcccccccccc@{}}
    \toprule
        \multirow{2}{*}[-2pt]{\textbf{$\mathcal{L}_{\text{D}}$}} &
        \multirow{2}{*}[-2pt]{\textbf{Method}} &
        \multicolumn{3}{c}{\textbf{Tanks~\&~Temples}} & 
        \multicolumn{3}{c}{\textbf{OMMO}}  & 
        \multicolumn{3}{c}{\textbf{DL3DV Benchmark}} \\
        \cmidrule(lr){3-5} \cmidrule(l){6-8} \cmidrule(l){9-11} 
         & &
        \textbf{PSNR}$\uparrow$ & \textbf{SSIM}$\uparrow$ & \textbf{LPIPS}$\downarrow$ & 
        \textbf{PSNR}$\uparrow$ & \textbf{SSIM}$\uparrow$ & \textbf{LPIPS}$\downarrow$ & 
        \textbf{PSNR}$\uparrow$ & \textbf{SSIM}$\uparrow$ & \textbf{LPIPS}$\downarrow$ 
        \\ 
    \midrule
        ~ & 3DGS & 
        23.73  & 0.853  & 0.169  & 
        30.49  & 0.920  & 0.142  & 
        27.16  & 0.860  & 0.167  \\ 
        \ding{55} & Seed Exp.   &
        \tidP{24.30}  & \tidP{0.861}  & 0.163  & 
        \tidP{30.85}  & 0.924  & \tidP{0.131}  & 
        27.74  & \tidP{0.869}  & \tidP{0.155}  \\ 
         ~ & Seed \& Split Exp. & 
        \sndP{24.37}  & \fstP{0.870}  & \fstP{0.139}  & 
        \fstP{31.27}  & \fstP{0.930}  & \fstP{0.119}  & 
        \fstP{28.43}  & \fstP{0.891}  & \fstP{0.124}  \\ 
        \midrule
        ~ & 3DGS &  
        24.06  & 0.856  & 0.166  & 
        30.59  & 0.923  & 0.136  & 
        27.33  & 0.864  & 0.162  \\ 
        \ding{51} & Seed Exp. &   
        24.19  & 0.859  & \tidP{0.162}  & 
        30.81  & \tidP{0.925}  & 0.132  & 
        \tidP{27.76}  & 0.865  & \tidP{0.155}  \\ 
        ~ & Seed \& Split Exp. &   
        \fstP{24.38}  & \sndP{0.869}  & \sndP{0.140}  & 
        \sndP{31.06}  & \sndP{0.928}  & \sndP{0.123}  & 
        \sndP{28.39}  & \sndP{0.888}  & \sndP{0.127} \\ 
    \bottomrule
  \end{tabular}
  }
  \caption{
    \textbf{Comparative Analysis: Deterministic Depth Supervision and Stochastic Exploration.}
    \(\mathcal{L}_{\text{D}}\)~(\(\mathcal{L}_{\text{Depth}}\)) integrates pseudo depth maps as a regularization term into the original loss function. 
    While incorporating \(\mathcal{L}_{\text{D}}\) into 3DGS yields consistent fidelity gains, it actively suppresses stochastic exploration dynamics. 
    Notably, sparse \textit{Random Seeding} alone delivers superior fidelity improvements compared to deterministic depth regularization, confirming that explicit exploration is more robust than depth priors for resolving depth ambiguities.
    }
  \label{tab:exp_depth_loss}
\end{table}

\subsection{Ablation Studies}
\label{ssec:ablation}

\paragraph{Random Seeding}
To validate the critical role of depth-directed signals, we conduct an experiment using heuristic depth supervision. 
We generate pseudo-depth maps for all input views using Depth Anything V2~\cite{yang2024depth} and integrate a depth regularization term ($\mathcal{L}_{\text{Depth}}$) into the standard 3DGS pipeline. Quantitative results across multiple unbounded scenes are summarized in \Table~\ref{tab:exp_depth_loss}. 

Incorporating $\mathcal{L}_{\text{Depth}}$ into the 3DGS consistently improves reconstruction fidelity, empirically confirming that the absence of depth-directed supervision is a primary driver of the Far-Side Blur Trap. 
However, pseudo-depth estimation is inherently imprecise. 
Its fidelity gains remain inferior to those achieved by sparse Random Seeding operator. More critically, integrating $\mathcal{L}_{\text{Depth}}$ into stochastic exploration yields diminishing returns. The deterministic exploitation of these inaccurate depth signals actively suppresses stochastic exploration dynamics, as the optimizer becomes constrained by unreliable priors.
As illustrated in \Figure~\ref{fig:depth_loss}, while integrating pseudo-depth supervision partially mitigates the Far-Side Blur, the inherent inaccuracy of pseudo-depth cues restricts its effective range. 
Consequently, distant regions remain vulnerable to persistent blurring as the optimizer becomes constrained by unreliable geometric priors. 
In contrast, \textit{Random Seeding} bypasses this limitation by explicitly probing uncharted depth intervals. 

These demonstrate that minimal random exploration provides a more robust and effective pathway to escape the Blur Trap than forcing gradient descent along heuristic depth regularization.
\textbf{\textit{An alternative depth exploration strategy is detailed in the \Appendix~\ref{ssec:pos_perturb_explor}, further underscoring the critical role of explicit exploration. }}

\paragraph{Random Splitting}

\begin{figure}[!t]
    \centering
    \includegraphics[trim={0.cm 0 0.cm 0},clip,width=0.97\linewidth]{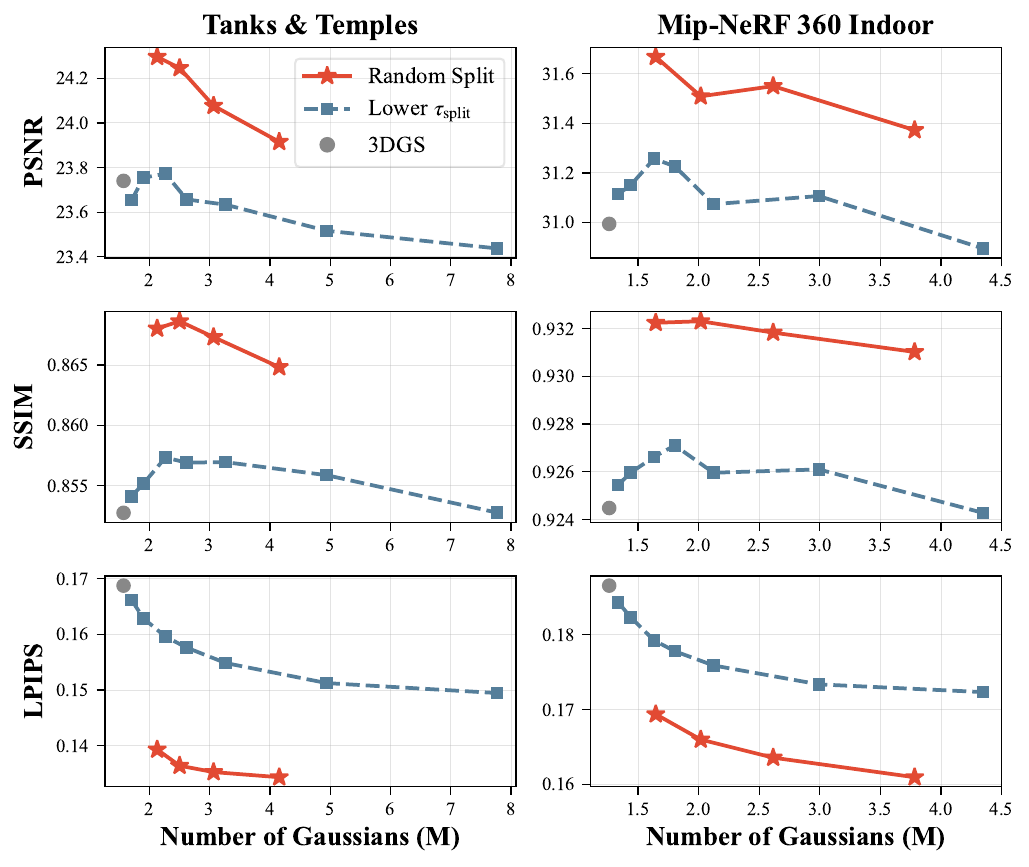}
    \caption{
    \textbf{Comparison of reconstruction fidelity gainst the number of Gaussian primitives.} 
    The dashed line represents standard 3DGS with lowered splitting threshold $\tau_{\text{split}}$, while solid line denote \textit{Random Splitting} at varying $N_{\text{split}}$. 
    Lower $\tau_{\text{split}}$ forces excessive splitting, leading to a sharp increase in primitive count. 
    \textit{Random Splitting} proactively targets large-scale Gaussians regardless of gradient suppression. 
    Sparse random splits outperforms exploitation-only splitting.
    }
    \label{fig:split_ablation}
\end{figure}

To verify that the fidelity gains from \textit{Random Splitting} do not merely stem from an increased Gaussian primitive count, we directly compare its efficiency against standard gradient-driven densification by evaluating the trade-off between primitive allocation and reconstruction fidelity. 
Specifically, we test \textit{Random Splitting} across varying intensities ($N_{\text{split}} \in \{20,50, 100, 200\}$). And we lower the 2D position gradient threshold $\tau_{\text{split}}$ on 3DGS. 
As shown in \Figure~\ref{fig:split_ablation}, simply reducing $\tau_{\text{split}}$ fails to yield meaningful fidelity improvements and instead introduces excessive redundant primitives. 
In contrast, \textbf{even a sparse application of \textit{Random Splitting} efficiently enhances reconstruction quality} while maintaining a significantly leaner primitive allocation. 
This confirms that the performance gain originates from targeted structural exploration rather than brute-force densification.

\section{Conclusion}
\label{sec:conclusion}

We identify the \textit{Blur Trap} as a fundamental optimization limitation in 3DGS, rooted in viewing-ray gradient orthogonality and blending-induced gradient attenuation that systematically suppress depth-directed signals. 
To break this exploitation-only deadlock, we introduce two minimal exploration operators: \textit{Random Seeding} and \textit{Random Splitting}. 
By bypassing gradient-dependent criteria, these operators proactively refine distant and occluded regions. 
Cross-benchmark evaluations confirm that their combination consistently escapes both Blur Trap, delivering state-of-the-art fidelity with negligible overhead and adaptive primitive allocation. 
This demonstrates that reconstruction bottlenecks stem from optimization dynamics, not model capacity. 
Our work establishes explicit exploration as a necessary component for differentiable rendering, advocating a shift toward balanced exploration-exploitation dynamics.

\clearpage

\clearpage
\bibliography{ref}               

\newpage
\appendix
\onecolumn
\begin{center}
	{\LARGE \bf Appendix}
\end{center}
\bigskip

\hrule
\vskip 0.2in
\startcontents[sections]
\printcontents[sections]{}{1}{\setcounter{tocdepth}{3}}
\vskip 0.2in
\hrule

\label{appendix}
\section{The Prove of Orthogonality}
\label{ssec:appendix_A}

\noindent\textbf{Goal.}
We prove that the 2D position path gradient $\mathbf{g}_{\mathrm{2d}} = \partial L^{2D} / \partial \mathbf{P}_{\mathrm{3D}}$ is always orthogonal to the viewing direction $\mathbf{D}_{\mathrm{C2G}} = \mathbf{P}_{\mathrm{3D}} - \mathbf{P}_{\mathrm{cam}}$. 
This reveals a fundamental limitation: the 2D position updates alone provides \textbf{no gradient along the line of sight}. 
The prove proceeds by decomposing $\mathbf{g}_{\mathrm{2d}}$ through the rendering pipeline via the chain rule and showing that the inner product $\mathbf{g}_{\mathrm{2d}} \cdot \mathbf{D}_{\mathrm{C2G}}$ collapses to zero through two geometric facts: (i)~$^{h}\mathbf{p}$ lies in the null space of the perspective division Jacobian, and (ii)~the viewport Jacobian annihilates the surviving $z$-component.

\subsection{Preliminaries}

\noindent \textbf{Symbol Definition}
To rigorously derive the orthogonality property, we first establish the notation used throughout this prove.
We denote the camera position in world coordinates as $\mathbf{P}_{\mathrm{cam}}$, with pose $[\mathbf{R},\mathbf{t}]$.
For a Gaussian primitive $\mathrm{G}$, let $\mathbf{P}_{\mathrm{3D}}$ be its world-space position and $\mathbf{p}_{\mathrm{cam}}$ its camera-space position.
The rendering pipeline transforms $\mathbf{P}_{\mathrm{3D}}$ through the following stages:
\begin{enumerate}
    \item The view matrix $\mathbf{V}$ maps world coordinates to camera space: $[\mathbf{p}_{\mathrm{cam}}^\mathrm{T}, 1]^\mathrm{T} = \mathbf{V} [\mathbf{P}_{\mathrm{3D}}^\mathrm{T}, 1]^\mathrm{T}$.
    \item The projection matrix $\mathbf{P}$ maps camera coordinates to homogeneous clip space: $^{h}\mathbf{p} = \mathbf{P} [\mathbf{p}_{\mathrm{cam}}^\mathrm{T}, 1]^\mathrm{T}$.
    \item Perspective division yields Normalized Device Coordinates (NDC): $\mathbf{p}_{\mathrm{ndc}} = [\frac{p_x}{p_w}, \frac{p_y}{p_w}, \frac{p_z}{p_w}]^\mathrm{T}$.
    \item The viewport transformation with image dimensions $W \times H$ produces 2D pixel coordinates $\mathbf{p}_{\mathrm{2D}}$.
\end{enumerate}
The composite matrix $\mathbf{M} = \mathbf{P} \mathbf{V}$ encodes the combined camera--projection mapping.
These symbols are summarized below.

\begin{equation}
\begin{aligned}
    \mathbf{P}_{\mathrm{3D}} &= [x,y,z]^\mathrm{T}  \\
    \mathbf{p}_{\mathrm{cam}} &= [p_x^{\mathrm{G}}, p_y^{\mathrm{G}}, p_z^{\mathrm{G}}]^\mathrm{T} \\
    ^{h}\mathbf{p} &= [p_x,p_y,p_z,p_w]^\mathrm{T} \\
    \mathbf{p}_{\mathrm{ndc}} &= [x_n,y_n,z_n]^\mathrm{T} \\
    \mathbf{p}_{\mathrm{2D}} &= [u,v]^\mathrm{T}  \\
    \mathbf{M} &= \mathbf{P} \mathbf{V}
    \\
    \mathbf{V} &= 
        \begin{bmatrix}
            R_{00}  &R_{01} &R_{02} &t_x \\
            R_{10}  &R_{11} &R_{12} &t_y \\
            R_{20}  &R_{21} &R_{22} &t_z \\
            0       &0      &0      &1   \\
        \end{bmatrix}
    \\
    \mathbf{P} &= 
            \begin{bmatrix}
                \frac{2n}{r-l}  &0                  &\frac{r+l}{r-l}    &0                  \\
                0               &\frac{2n}{t-b}     &\frac{t+b}{t-b}    &0                  \\
                0               &0                  &-\frac{f+n}{f-n}   &-\frac{2fn}{f-n}   \\
                0               &0                  &-1                 &0                  \\
            \end{bmatrix}
\label{eq:the_point}
\end{aligned}
\end{equation}

\noindent \textbf{Forward Mapping}
The forward mapping proceeds through four stages:

\textit{World to camera.}

\begin{equation}
\begin{aligned}
    \begin{bmatrix}
        \mathbf{p}_{\mathrm{cam}}^\mathrm{T}, 1
    \end{bmatrix}^\mathrm{T}
    &=
    \begin{bmatrix}
        p_x^{\mathrm{G}}, p_y^{\mathrm{G}}, p_z^{\mathrm{G}}, 1
    \end{bmatrix}^\mathrm{T}
    = \mathbf{V}
    \begin{bmatrix}
        \mathbf{P}_{\mathrm{3D}}^\mathrm{T}, 1
    \end{bmatrix}^\mathrm{T},
    \label{eq:forward_w2c}
\end{aligned}
\end{equation}

\textit{Camera to clip.}

\begin{equation}
\begin{aligned}
    ^{h}\mathbf{p}
    &=
    \begin{bmatrix}
        p_x, p_y, p_z, p_w
    \end{bmatrix}^\mathrm{T}
    = \mathbf{M}
    \begin{bmatrix}
        \mathbf{P}_{\mathrm{3D}}^\mathrm{T}, 1
    \end{bmatrix}^\mathrm{T}
    = \mathbf{P}
    \begin{bmatrix}
        \mathbf{p}_{\mathrm{cam}}^\mathrm{T}, 1
    \end{bmatrix}^\mathrm{T},
    \label{eq:forward_c2clip}
\end{aligned}
\end{equation}

\textit{Clip to NDC.}

\begin{equation}
    \mathbf{p}_{\mathrm{ndc}} =
    \begin{bmatrix}
        x_n \\[2pt] y_n \\[2pt] z_n
    \end{bmatrix}
    =
    \begin{bmatrix}
        p_x/p_w \\[2pt] p_y/p_w \\[2pt] p_z/p_w
    \end{bmatrix},
    \label{eq:forward_clip2ndc}
\end{equation}

\textit{NDC to pixel.}

\begin{equation}
    \mathbf{p}_{\mathrm{2D}} =
    \begin{bmatrix}
        u \\[2pt] v
    \end{bmatrix}
    =
    \begin{bmatrix}
        (x_n+1)\cdot\frac{W}{2} \\[4pt]
        (y_n+1)\cdot\frac{H}{2}
    \end{bmatrix}.
    \label{eq:forward_ndc2pix}
\end{equation}

\subsection{Gradient Decomposition}

The gradient of the loss with respect to the 3D Gaussian position comprises three terms:
\begin{equation}
\begin{aligned}
    \frac{\partial L}{\partial \mathbf{P}_{\mathrm{3D}}}
        &=
        \frac{\partial L}{\partial \mathbf{Cov2D}} \frac{\partial \mathbf{Cov2D}}{\partial \mathbf{P}_{\mathrm{3D}}}
        +
        \frac{\partial L}{\partial \mathbf{p}_{\mathrm{2D}}} \frac{\partial \mathbf{p}_{\mathrm{2D}}}{\partial \mathbf{P}_{\mathrm{3D}}}
        +
        \frac{\partial L}{\partial \mathbf{SH}} \frac{\partial \mathbf{SH}}{\partial \mathbf{P}_{\mathrm{3D}}}.
        \label{eq:dL_dP3D_total}
\end{aligned}
\end{equation}
These correspond to the optimization paths through the 2D covariance, the 2D screen-space position, and the spherical harmonics coefficients, respectively.
Among them, the 2D position term dominates in magnitude; we therefore focus on this path and decompose it via the chain rule into three Jacobian matrices.

The 2D position gradient further decomposes via the chain rule through the rendering pipeline:
\begin{equation}
\begin{aligned}
    \mathbf{g}_{\mathrm{2d}}
        =
        \frac{\partial L^{2D}}{\partial \mathbf{P}_{\mathrm{3D}}}
        &=
        \frac{\partial L}{\partial \mathbf{p}_{\mathrm{2D}}}
        \frac{\partial \mathbf{p}_{\mathrm{2D}}}{\partial \mathbf{p}_{\mathrm{ndc}}}
        \frac{\partial \mathbf{p}_{\mathrm{ndc}}}{\partial ^{h}\mathbf{p}}
        \frac{\partial ^{h}\mathbf{p}}{\partial \mathbf{P}_{\mathrm{3D}}}
        \label{eq:dL_d2D_d3D}
\end{aligned}
\end{equation}
The core factor $\frac{\partial \mathbf{p}_{\mathrm{2D}}}{\partial \mathbf{P}_{\mathrm{3D}}}$ is the product of three Jacobian matrices, which we now derive explicitly.

\begin{equation}
    \frac{\partial \mathbf{p}_{\mathrm{2D}}}{\partial \mathbf{p}_{\mathrm{ndc}}} 
        =
        \begin{bmatrix}
            \frac{\partial u}{\partial x_n} &\frac{\partial u}{\partial y_n} &\frac{\partial u}{\partial z_n} \\
            \frac{\partial v}{\partial x_n} &\frac{\partial v}{\partial y_n} &\frac{\partial v}{\partial z_n}
        \end{bmatrix}
        =
        \begin{bmatrix}
            \frac{W}{2} &0 &0 \\
            0 &\frac{H}{2} &0
        \end{bmatrix}.
    \label{eq:jac_viewport}
\end{equation}

\begin{equation}
    \begin{aligned}
    \frac{\partial \mathbf{p}_{\mathrm{ndc}}}{\partial ^{h}\mathbf{p}}
        &=
        \begin{bmatrix}
            \frac{\partial x_n}{\partial p_x} &\frac{\partial x_n}{\partial p_y} &\frac{\partial x_n}{\partial p_z} &\frac{\partial x_n}{\partial p_w} \\
            \frac{\partial y_n}{\partial p_x} &\frac{\partial y_n}{\partial p_y} &\frac{\partial y_n}{\partial p_z} &\frac{\partial y_n}{\partial p_w} \\
            \frac{\partial z_n}{\partial p_x} &\frac{\partial z_n}{\partial p_y} &\frac{\partial z_n}{\partial p_z} &\frac{\partial z_n}{\partial p_w} 
        \end{bmatrix}   \\
        &=
        \begin{bmatrix}
            \frac{1}{p_w} &0 &0 &-\frac{p_x}{p_w^2} \\
            0 &\frac{1}{p_w} &0 &-\frac{p_y}{p_w^2} \\
            0 &0 &\frac{1}{p_w} &-\frac{p_z}{p_w^2} 
        \end{bmatrix}.
    \label{eq:jac_persp}
    \end{aligned}
\end{equation}

\begin{equation}
    \begin{aligned}
    \frac{\partial ^{h}\mathbf{p}}{\partial \mathbf{P}_{\mathrm{3D}}}
        &=
        \begin{bmatrix}
            \frac{\partial p_x}{\partial x} &\frac{\partial p_x}{\partial y} &\frac{\partial p_x}{\partial z} \\
            \frac{\partial p_y}{\partial x} &\frac{\partial p_y}{\partial y} &\frac{\partial p_y}{\partial z} \\
            \frac{\partial p_z}{\partial x} &\frac{\partial p_z}{\partial y} &\frac{\partial p_z}{\partial z} \\
            \frac{\partial p_w}{\partial x} &\frac{\partial p_w}{\partial y} &\frac{\partial p_w}{\partial z} 
        \end{bmatrix}   \\
        &=
        \begin{bmatrix}
            M_{00} &M_{01} &M_{02} \\
            M_{10} &M_{11} &M_{12} \\
            M_{20} &M_{21} &M_{22} \\
            M_{30} &M_{31} &M_{32}
        \end{bmatrix}   \\
        &=
        \mathbf{M}_{[:,:3]} = \mathbf{P} \mathbf{V_{[:, :3]}}.
    \label{eq:jac_proj}
    \end{aligned}
\end{equation}

For the orthogonality prove we also need the world-space direction from the camera to the Gaussian, $\mathbf{P}_{\mathrm{3D}} - \mathbf{P}_{\mathrm{cam}}$.
Applying $\mathbf{V}_{[:,:3]}$ to this direction and then $\mathbf{P}$ yields:
\begin{equation}
    \mathbf{P} \mathbf{V_{[:, :3]}}
        (\mathbf{P}_{\mathrm{3D}} - \mathbf{P}_{\mathrm{cam}})
    = \mathbf{P}
        \begin{bmatrix}
            \mathbf{p}_{\mathrm{cam}}^\mathrm{T}, 0
        \end{bmatrix}^\mathrm{T},
    \label{eq:vp}
\end{equation}
which, after perspective division, gives the vanishing point of all lines parallel to the camera-to-Gaussian ray $\mathbf{p}_{\mathrm{cam}}$ in camera space.

\subsection{Prove of Orthogonality}

We prove that $\mathbf{g}_{\mathrm{2d}}$ is always orthogonal to the viewing direction from the camera center to the primitive.
This orthogonality property is formalized in~\Equation~\ref{eq:dL_d2D_d3D_final}.

Let $\mathbf{D}_{\mathrm{C2G}} = \mathbf{P}_{\mathrm{3D}} - \mathbf{P}_{\mathrm{cam}}$ be the viewing direction from the camera to the Gaussian.
Using the gradient decomposition from~\Equation~\ref{eq:dL_d2D_d3D}, their inner product expands as
\begin{equation}
\begin{aligned}
    \frac{\partial L^{2D}}{\partial \mathbf{P}_{\mathrm{3D}}} \mathbf{D}_{\mathrm{C2G}} 
        &=
        \frac{\partial L^{2D}}{\partial \mathbf{P}_{\mathrm{3D}}} (\mathbf{P}_{\mathrm{3D}} - \mathbf{P}_{\mathrm{cam}})
        \\
        &= 
        \frac{\partial L}{\partial \mathbf{p}_{\mathrm{2D}}} 
        \frac{\partial \mathbf{p}_{\mathrm{2D}}}{\partial \mathbf{p}_{\mathrm{ndc}}}
        \frac{\partial \mathbf{p}_{\mathrm{ndc}}}{\partial ^{h}\mathbf{p}} 
        \frac{\partial ^{h}\mathbf{p}}{\partial \mathbf{P}_{\mathrm{3D}}}
        (\mathbf{P}_{\mathrm{3D}} - \mathbf{P}_{\mathrm{cam}})    \\
        &=
        \frac{\partial L}{\partial \mathbf{p}_{\mathrm{2D}}} 
        \frac{\partial \mathbf{p}_{\mathrm{2D}}}{\partial \mathbf{p}_{\mathrm{ndc}}}
        \frac{\partial \mathbf{p}_{\mathrm{ndc}}}{\partial ^{h}\mathbf{p}} 
         \mathbf{P} \mathbf{V_{[:, :3]}} 
         (\mathbf{P}_{\mathrm{3D}} - \mathbf{P}_{\mathrm{cam}})
         \\
         &= 
         \frac{\partial L}{\partial \mathbf{p}_{\mathrm{2D}}} 
        \frac{\partial \mathbf{p}_{\mathrm{2D}}}{\partial \mathbf{p}_{\mathrm{ndc}}}
        \frac{\partial \mathbf{p}_{\mathrm{ndc}}}{\partial ^{h}\mathbf{p}} 
         \mathbf{P}  
            \begin{bmatrix}
                \mathbf{p}_{\mathrm{cam}}^\mathrm{T}, 0
            \end{bmatrix}^\mathrm{T}
        \\
        &=
        \frac{\partial L}{\partial \mathbf{p}_{\mathrm{2D}}} 
        \frac{\partial \mathbf{p}_{\mathrm{2D}}}{\partial \mathbf{p}_{\mathrm{ndc}}}
        \frac{\partial \mathbf{p}_{\mathrm{ndc}}}{\partial ^{h}\mathbf{p}} 
            \begin{bmatrix}
                \frac{2n}{r-l}  &0                  &\frac{r+l}{r-l}    &0                  \\
                0               &\frac{2n}{t-b}     &\frac{t+b}{t-b}    &0                  \\
                0               &0                  &-\frac{f+n}{f-n}   &-\frac{2fn}{f-n}   \\
                0               &0                  &-1                 &0                  \\
            \end{bmatrix}
            \begin{bmatrix}
                p^{\mathrm{G}}_x  \\
                p^{\mathrm{G}}_y  \\
                p^{\mathrm{G}}_z  \\
                0
            \end{bmatrix}
        \\
        &=
        \frac{\partial L}{\partial \mathbf{p}_{\mathrm{2D}}} 
        \frac{\partial \mathbf{p}_{\mathrm{2D}}}{\partial \mathbf{p}_{\mathrm{ndc}}}
        \frac{\partial \mathbf{p}_{\mathrm{ndc}}}{\partial ^{h}\mathbf{p}} 
            \begin{bmatrix}
                \frac{2n}{r-l} p^{\mathrm{G}}_x + \frac{r+l}{r-l} p^{\mathrm{G}}_z  \\
                \frac{2n}{t-b} p^{\mathrm{G}}_y + \frac{t+b}{t-b}p^{\mathrm{G}}_z    \\
                -\frac{f+n}{f-n} p^{\mathrm{G}}_z \\
                -p^{\mathrm{G}}_z\\
            \end{bmatrix}
        \\
        &=
        \frac{\partial L}{\partial \mathbf{p}_{\mathrm{2D}}} 
        \frac{\partial \mathbf{p}_{\mathrm{2D}}}{\partial \mathbf{p}_{\mathrm{ndc}}}
        \frac{\partial \mathbf{p}_{\mathrm{ndc}}}{\partial ^{h}\mathbf{p}} 
            \begin{bmatrix}
                p_x\\
                p_y\\
                p_z + \frac{2fn}{f-n}\\
                p_w\\
            \end{bmatrix}
        \\
        &=
        \frac{\partial L}{\partial \mathbf{p}_{\mathrm{2D}}} 
        \frac{\partial \mathbf{p}_{\mathrm{2D}}}{\partial \mathbf{p}_{\mathrm{ndc}}}
        \begin{bmatrix}
            \frac{1}{p_w} &0 &0 &-\frac{p_x}{p_w^2} \\
            0 &\frac{1}{p_w} &0 &-\frac{p_y}{p_w^2} \\
            0 &0 &\frac{1}{p_w} &-\frac{p_z}{p_w^2} 
        \end{bmatrix}
        \begin{bmatrix}
            p_x\\
            p_y\\
            p_z + \frac{2fn}{f-n}\\
            p_w\\
        \end{bmatrix}
        \\
        &=
        \frac{\partial L}{\partial \mathbf{p}_{\mathrm{2D}}} 
        \begin{bmatrix}
            \frac{W}{2} &0 &0 \\
            0 &\frac{H}{2} &0
        \end{bmatrix}
        \begin{bmatrix}
            0   \\
            0   \\
            \frac{2fn}{(f-n)p_w}
        \end{bmatrix}
        \\
        &=
        \frac{\partial L}{\partial \mathbf{p}_{\mathrm{2D}}} 
        \begin{bmatrix}
            0 \\
            0 
        \end{bmatrix}
        \\
        &=
        0
    \end{aligned}
\end{equation}
\begin{equation}
\begin{aligned}
    \frac{\partial L^{2D}}{\partial \mathbf{P}_{\mathrm{3D}}} \mathbf{D}_{\mathrm{C2G}}
        &=
        \frac{\partial L^{2D}}{\partial \mathbf{P}_{\mathrm{3D}}} (\mathbf{P}_{\mathrm{3D}} - \mathbf{P}_{\mathrm{cam}})
        \label{eq:dL_d2D_d3D_final}
        =
        0
\end{aligned}
\end{equation}

The derivation above reveals two geometric reasons behind this orthogonality.
First, since $\mathbf{p}_{\mathrm{ndc}}$ is homogeneous of degree zero in $^{h}\mathbf{p}$,
the homogeneous coordinate vector $^{h}\mathbf{p}$ lies in the null space of $\frac{\partial \mathbf{p}_{\mathrm{ndc}}}{\partial ^{h}\mathbf{p}}$.
Combined with $\mathbf{P} [\mathbf{p}_{\mathrm{cam}}^\mathrm{T}, 0]^\mathrm{T} = {^{h}\mathbf{p}} + [0,\,0,\,\frac{2fn}{f-n},\,0]^\mathrm{T}$,
this forces the intermediate vector after perspective division to carry only a $z$-component, i.e., $\bigl[0,\;0,\;\frac{2fn}{(f-n)p_w}\bigr]^\mathrm{T}$.
Second, left-multiplying by the viewport Jacobian $\frac{\partial \mathbf{p}_{\mathrm{2D}}}{\partial \mathbf{p}_{\mathrm{ndc}}}$, whose third column is identically zero, annihilates this $z$-component and yields $[0,\;0]^\mathrm{T}$.
Multiplying by $\frac{\partial L}{\partial \mathbf{p}_{\mathrm{2D}}}$ then trivially gives zero.
Consequently, $\mathbf{D}_{\mathrm{C2G}}$ is always orthogonal to $\frac{\partial L^{2D}}{\partial \mathbf{P}_{\mathrm{3D}}}$.
\myblue{In essence, the orthogonality arises from two geometric facts: $^{h}\mathbf{p}$ lies in the null space of the perspective division Jacobian, forcing the intermediate vector to a pure $z$-component; the viewport Jacobian then annihilates this component.}
\hfill $\Box$

\section{Factor-wise Dependence of the 2D Positional Gradient on Depth Sorting}
\label{ssec:appendix_B}

\noindent\textbf{Goal.}
We prove that along a pixel's ray, the 2D positional gradient magnitude $|\partial L / \partial x_m|$ of a depth-sorted Gaussian $\mathrm{G}_m$ decays rapidly as the sorted index $m$~increases, making rear Gaussians increasingly unable to surpass the densification threshold $\tau$.
This decay is driven by the accumulated transmittance $T_{m-1} = \prod_{j=1}^{m-1}(1-\alpha_j)$ of all front Gaussians: $T_{m-1}$ is monotonically non-increasing in~$m$, so the gradient bound compresses toward zero---the direct cause of \textbf{blending-induced densification failure}. We formalize this by decomposing the gradient into factors, identifying the sole $m$-dependent factor, and bounding its magnitude by $T_{m-1}$.

\subsection{Gradient Decomposition}
\label{sec:C_gradient_formulation}

For a pixel $p$, let the depth-sorted Gaussians along its ray be $\mathrm{G}_1, \dots, \mathrm{G}_N$.
The rendered color follows the \(\alpha\)-blending equation:
\begin{equation}
\begin{aligned}
    \hat{\mathbf{c}}^{(p)} &= \sum_{i=1}^{N} \mathbf{c}_i \alpha_i^{(p)} T_{i-1}^{(p)} + \mathbf{c}_{\mathrm{bg}}\,
    T_N^{(p)},
    \\[2pt]
    T_k^{(p)} &= \prod_{j=1}^{k} (1-\alpha_j^{(p)}),
\label{eq:C_blending}
\end{aligned}
\end{equation}
where $\alpha_i^{(p)} = o_i \cdot G_i^{\mathrm{2D}}(p)$ is the effective opacity of $\mathrm{G}_i$ at pixel~$p$, with learned opacity $o_i \in (0,1)$ and 2D Gaussian $G_i^{\mathrm{2D}}(p) \in (0,1]$; colors satisfy $\mathbf{c}_i, \mathbf{c}_{\mathrm{bg}} \in [0,1]^3$.

We decompose $\partial L / \partial x_m$ via the chain rule:
\begin{equation}
    \frac{\partial L}{\partial x_m} =
    \sum_{p} \sum_{c=1}^{C}
    \underbrace{\frac{\partial L}{\partial \hat{\mathbf{c}}_c^{(p)}}}_{\text{Factor A}}
    \cdot
    \underbrace{\frac{\partial \hat{\mathbf{c}}_c^{(p)}}{\partial \alpha_m^{(p)}}}_{\text{Factor B}}
    \cdot
    \underbrace{o_m}_{\text{Factor C}}
    \cdot
    \underbrace{\frac{\partial G_m^{\mathrm{2D},(p)}}{\partial x_m}}_{\text{Factor D}},
    \label{eq:C_dL_dx}
\end{equation}
and identically for $\partial L / \partial y_m$ with $\partial G_m^{\mathrm{2D},(p)} / \partial y_m$ in Factor~D.

\subsection{$m$-Dependence of Each Factor}
\label{sec:C_factor_analysis}

We examine which factors in~\Equation~\ref{eq:C_dL_dx} are influenced by the depth-sorted index~$m$.

\noindent\textbf{Factor A} \;($\partial L / \partial \hat{\mathbf{c}}_c^{(p)}$): \emph{independent of $m$}.
This term is the derivative of the loss with respect to the rendered pixel color; it does not involve~$m$ and is therefore independent of depth ordering.

\noindent\textbf{Factor C} \;($o_m$): \emph{independent of $m$}.
The opacity is a learned intrinsic parameter of Gaussian~$\mathrm{G}_m$. Its value is attached to the primitive itself and does not change when the view-dependent sorting reassigns which Gaussian occupies slot~$m$.

\noindent\textbf{Factor D} \;($\partial G_m^{\mathrm{2D},(p)} / \partial x_m$): \emph{independent of $m$}.
This spatial derivative of the 2D Gaussian kernel depends only on the screen-space covariance $\Sigma_m$ and the pixel offset $\boldsymbol{\Delta}^{(p)} = [x_m - x_p,\, y_m - y_p]^{\!\top}$---quantities determined by the intrinsic 3D position and covariance of $\mathrm{G}_m$, neither of which is affected by the depth-sorted ordering.

\noindent\textbf{Factor B} \;($\partial \hat{\mathbf{c}}_c^{(p)} / \partial \alpha_m^{(p)}$): \emph{depends on $m$}.
This factor couples $\mathrm{G}_m$ to every other Gaussian through the \(\alpha\)-blending equation. It is the sole channel through which the depth-sorted index~$m$ enters the gradient. We now derive its explicit form.

\subsection{Derivation of Factor B $\partial \hat{\mathbf{c}}_c^{(p)} / \partial \alpha_m^{(p)}$}
\label{sec:C_factor_b_derivation}

Dropping the pixel superscript $(p)$ for brevity, we split~\Equation~\ref{eq:C_blending} at index~$m$ to isolate terms involving $\alpha_m$:
\begin{equation}
    \hat{\mathbf{c}} =
    \underbrace{\sum_{i=1}^{m-1} \mathbf{c}_i \alpha_i T_{i-1}}_{\text{in front of } \mathrm{G}_m}
    \;+\;
    \underbrace{\mathbf{c}_m \alpha_m T_{m-1}}_{\mathrm{G}_m \text{ itself}}
    \;+\;
    \underbrace{\sum_{i=m+1}^{N} \mathbf{c}_i \alpha_i T_{i-1} + \mathbf{c}_{\mathrm{bg}} T_N}_{\text{behind } \mathrm{G}_m}.
    \label{eq:C_split}
\end{equation}

Differentiating~\Equation~\ref{eq:C_split} w.r.t.\ $\alpha_m$~\cite{wang2025faster}:
\begin{equation}
\begin{aligned}
    \frac{\partial \hat{\mathbf{c}}}{\partial \alpha_m}
    &= T_{m-1} \Biggl[ \mathbf{c}_m
       \;-\; \underbrace{\Biggl( \sum_{i=m+1}^{N} \mathbf{c}_i \alpha_i \!\!\prod_{j=m+1}^{i-1} (1-\alpha_j)
       + \mathbf{c}_{\mathrm{bg}} \!\!\prod_{j=m+1}^{N} (1-\alpha_j) \Biggr)}_{\displaystyle =\, \hat{\mathbf{c}}_{m+1:N} \,\in\, [0,1]^3}
       \Biggr]
    \\[6pt]
    &= T_{m-1} \cdot \bigl( \mathbf{c}_m - \hat{\mathbf{c}}_{m+1:N} \bigr),
\label{eq:C_dC_dalpha_derived}
\end{aligned}
\end{equation}
where $\hat{\mathbf{c}}_{m+1:N}$ denotes the alpha-composited color of the Gaussians behind $\mathrm{G}_m$ over the background.

Since both $\mathbf{c}_m$ and $\hat{\mathbf{c}}_{m+1:N}$ lie in $[0,1]^3$, their per-channel difference is bounded by~$1$, yielding the bound
\begin{equation}
    \left| \frac{\partial \hat{\mathbf{c}}_c}{\partial \alpha_m} \right|
    \leq T_{m-1},
    \quad \forall\,c \in \{1,2,3\},
    \label{eq:C_factor_b_bound}
\end{equation}
where $T_{m-1} = \prod_{j=1}^{m-1}(1 - \alpha_j)$ is the accumulated transmittance of the $m-1$ Gaussians in front of $\mathrm{G}_m$.

\medskip
\noindent\textbf{Monotonicity of the bound.}
Each $\alpha_j \in [0,1]$ implies $1 - \alpha_j \in [0,1]$, hence $T_k = T_{k-1} \cdot (1 - \alpha_k) \leq T_{k-1}$ for all $k \geq 1$.
The transmittance sequence is therefore monotonically non-increasing:
\begin{equation}
    T_0 = 1 \;\geq\; T_1 \;\geq\; T_2 \;\geq\; \cdots \;\geq\; T_N \;\geq\; 0.
\end{equation}
Consequently, the upper bounds in~\Equation~\ref{eq:C_factor_b_bound} can only decrease or stay constant as the depth-sorted index~$m$ grows.

\medskip

\subsection{Gradient attenuation}

The exact gradient $T_{m-1} \cdot (\mathbf{c}_m - \hat{\mathbf{c}}_{m+1:N})$ depends on $m$ through two factors: the transmittance $T_{m-1}$, which decays monotonically, and the color difference $\mathbf{c}_m - \hat{\mathbf{c}}_{m+1:N}$. The latter varies non-monotonically with~$m$, because advancing $m$ alters both which Gaussian occupies slot~$m$ and the rear subsequence $\hat{\mathbf{c}}_{m+1:N}$ being composited.

While the color-difference term may cause the gradient magnitude to fluctuate, the magnitude can never exceed the monotonic envelope $T_{m-1}$, and this envelope decays monotonically toward zero as front Gaussians accumulate.
This attenuation is inescapable: $T_{m-1}$ depends only on the Gaussians in front of position~$m$, so the gradient bound lies entirely outside the control of rear Gaussians.

As $m$~increases, rear Gaussians therefore receive a progressively weaker 2D positional gradient signal, making them increasingly unable to surpass the densification threshold $\tau$---the direct mechanism of blending-induced densification failure. 
\hfill $\square$

\section{The Positional Perturbation Exploration}

\label{ssec:pos_perturb_explor}

\begin{figure}[!ht]
    \centering
    \includegraphics[trim={0.cm 0 0.cm 0},clip,width=0.99\linewidth]{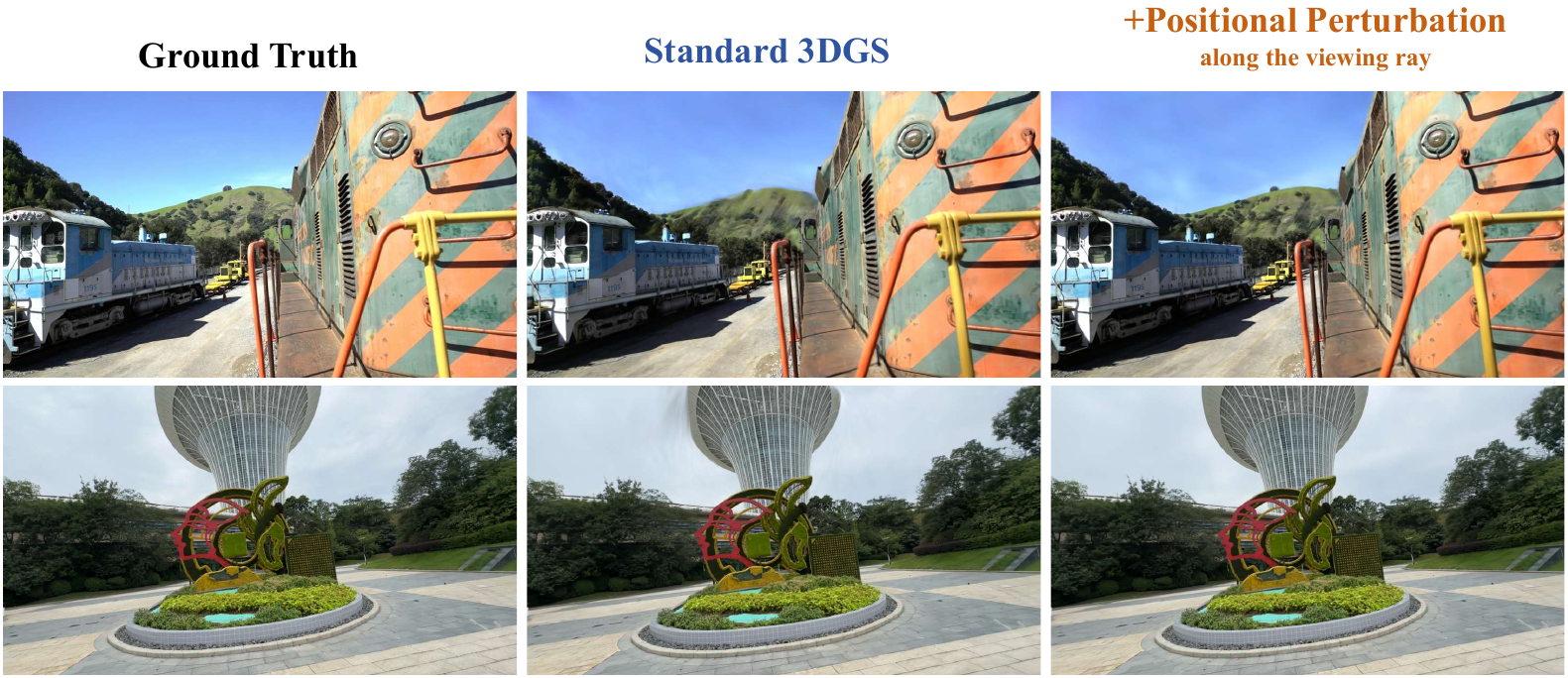}
    \caption{\textbf{Positional perturbation exploration along the viewing ray.}
    }
    \label{fig:pos_pert}
\end{figure}

We adapt the positional perturbation mechanism~\cite{kheradmand20243d} from 3DGS-MCMC to the standard 3DGS pipeline, constraining perturbation direction strictly along the viewing ray. The results demonstrate that even this straightforward depth-directed intervention significantly enhances reconstruction fidelity in unbounded scenes, as shown in ~\Figure~\ref{fig:pos_pert}.

\end{document}